\PassOptionsToPackage{dvipsnames}{xcolor}
\documentclass[11pt,letterpaper]{style}

\usepackage[numbers]{natbib}
\usepackage{graphicx}
\usepackage{booktabs}
\usepackage{amsmath,amsfonts,amssymb}
\usepackage{cleveref}
\usepackage{subcaption}
\usepackage{wrapfig}
\usepackage{multirow}
\usepackage{colortbl}
\usepackage{listings}
\usepackage{xparse}
\usepackage{fontawesome5}
\usepackage{bxcoloremoji}
\usepackage{float}
\usepackage{placeins}
\usepackage{threeparttable}

\graphicspath{{./}{fig/}{figures/}{plot/}{pdf/}{table/}}
\usepackage{amsthm}
\usepackage{tcolorbox}
\usepackage{svg}
\tcbuselibrary{skins,breakable}
\tcbuselibrary{listingsutf8}
\usepackage{titletoc}

\usepackage{setspace}
\usepackage{pifont}
\usepackage{mathtools}
\usepackage{enumitem}
\usepackage{arydshln}
\usepackage{bbm}
\usepackage{lineno}
\usepackage{makecell}
\usepackage{adjustbox}
\usepackage{algorithm}
\usepackage{algorithmic}
\usepackage{caption} % 为了 \captionof{figure}{...}
\usepackage{wrapfig}

\definecolor{mygray}{gray}{0.9}
\definecolor{syncol}{RGB}{243,246,249}
\definecolor{wildcol}{RGB}{215,240,235}
\definecolor{drop1}{RGB}{180,225,220}
\definecolor{drop2}{RGB}{150,210,200}
\definecolor{drop3}{RGB}{120,195,185}
\definecolor{drop4}{RGB}{95,180,170}
\definecolor{drop5}{RGB}{65,160,150}
\definecolor{lightblue}{RGB}{26,82,249}

\definecolor{myblue1}{HTML}{0171DC}
\definecolor{myblue2}{HTML}{013978}

\NewDocumentEnvironment{minted}{O{} m +b}{%
}{}

\newcommand{\equal}{\textsuperscript{*}}     % Equal contribution
\newcommand{\advisor}{\textsuperscript{\dag}}% Equal advising

\renewcommand\Authfont{\centering\normalfont\bfseries\fontsize{11}{15}\selectfont}
\renewcommand\Affilfont{\centering\normalfont\fontsize{10}{15}\selectfont}

\title{Group-Evolving Agents: Open-Ended Self-Improvement via Experience Sharing}
\runningtitle{Group-Evolving Agents: Open-Ended Self-Improvement via Experience Sharing}

\author{%
    {\Authfont
    \textbf{Zhaotian Weng}\textsuperscript{1} \quad
    \textbf{Antonis Antoniades}\textsuperscript{1} \\ 
    \textbf{Deepak Nathani}\textsuperscript{1} \quad
    \textbf{Zhen Zhang}\textsuperscript{1} \quad
    \textbf{Xiao Pu} \textsuperscript{1} \quad
    \textbf{Xin Eric Wang}\advisor\textsuperscript{1}
    }\\
    {\Affilfont
    \textsuperscript{1}  University of California, Santa Barbara \quad \\
     \texttt{\{zhaotian,ericxwang\}@ucsb.edu}
   
    }
}

\begin{document}
\begin{abstract}
Open-ended self-improving agents can autonomously modify their own structural designs to advance their capabilities and overcome the limits of pre-defined architectures, thus reducing reliance on human intervention. We introduce \textbf{Group-Evolving Agents (GEA)}, a new paradigm for open-ended self-improvements, which treats a group of agents as the fundamental evolutionary unit, enabling explicit experience sharing and reuse within the group throughout evolution. Unlike existing open-ended self-evolving paradigms that adopt tree-structured evolution, GEA overcomes the limitation of inefficient utilization of exploratory diversity caused by isolated evolutionary branches. We evaluate GEA on challenging coding benchmarks, where it significantly outperforms state-of-the-art self-evolving methods (71.0\% vs. 56.7\% on SWE-bench Verified, 88.3\% vs. 68.3\% on Polyglot) and matches or exceeds top human-designed agent frameworks (71.8\% and 52.0\% on two benchmarks, respectively). Analysis reveals that GEA more effectively converts early-stage exploratory diversity into sustained, long-term progress, achieving stronger performance under the same number of evolved agents. Furthermore, GEA exhibits consistent transferability across different coding models and greater robustness, fixing framework-level bugs in 1.4 iterations on average, versus 5 for self-evolving methods.
\end{abstract}
\newcommand{\TitleLinks}{%
\centering
    \vspace{8pt}
    % {\noindent\absfont\fontsize{11}{13}\selectfont
    % \faGithub\ Project Page: \url{}\par}%
}
\maketitle

\begin{figure}[htbp]
    \centering
    \includegraphics[width=0.9\textwidth]{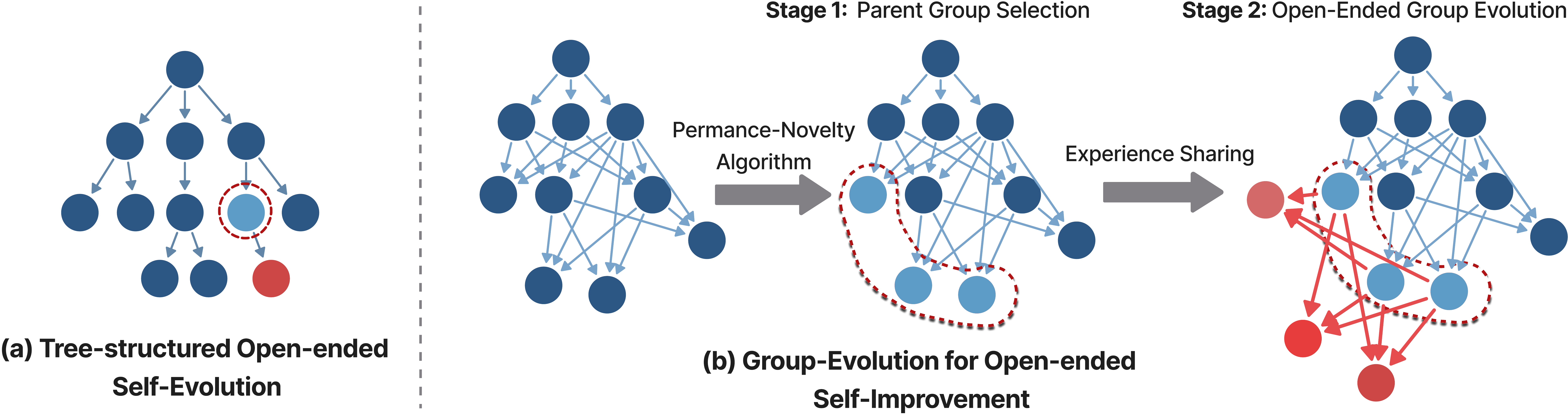}
    \caption{\textbf{Overview of Group-Evolving Agents (GEA) vs.\ tree-structured self-evolution for open-endedness.}
    GEA treats a \emph{group of agents}, rather than an individual agent, as the fundamental unit of evolution. At each iteration, a parent group jointly gives rise to an offspring group through explicit intra-group Experience sharing and reuse.}
    \label{fig:pipeline-overall}
\end{figure}

\section{Introduction}

Open-endedness and cumulative progress are key characteristics of scientific breakthroughs~\citep{stanley2017open,zhang2025darwin,stanley2015greatness}. However, most existing AI systems rely on pre-defined model architectures designed by humans. Although such systems can accumulate experience through training, they often struggle to transcend the capability boundaries imposed by their initial designs, as they lack the ability to modify their own structural configurations~\citep{yin2025godel}. Thus, progress remains heavily dependent on continuous human intervention.

Existing open-ended self-improving systems are largely inspired by biological evolution and designed around individual-centric evolutionary processes~\citep{zhang2025darwin,yin2025godel,schmidhuber2003godel,robeyns2025self,novikov2025alphaevolve}. At each iteration, a single agent is selected as the parent and refined to produce one or more offspring(Figure~\ref{fig:pipeline-overall}a). The overall structure follows chain- or tree-structured evolution, where different branches remain strictly isolated. Consequently, although such systems often exhibit substantial exploratory diversity, this diversity rarely serves as effective stepping stones~\citep{mouret2015illuminating, pugh2016quality}. Instead, many agents provide only temporary diversity, producing short-lived variants that fail to contribute to long-term cumulative progress.

It is time to rethink agent evolution. \emph{AI agents are not biological individuals; why should their evolution remain constrained by biological paradigms?}
In fact, AI agents can directly share trajectories, tools, and learned artifacts, and they can aggregate complementary skills without the constraints of reproduction or lineage.

Therefore, we introduce \textbf{Group-Evolving Agents (GEA)}, a new paradigm for open-ended self-improvement that treats a \emph{group of agents}, rather than an individual agent, as the fundamental unit of evolution (Figure~\ref{fig:pipeline-overall}b). 
This shift enables explicit experience sharing and reuse across agents within a group, naturally allowing exploratory discoveries from different agents to be consolidated and accumulated into long-term progress rather than remaining as short-lived variants.
At each iteration, GEA first selects a parent group of agents using a Performance-Novelty criterion that balances immediate performance gains with evolutionary diversity. The parent agents then jointly produce a child group through a shared pool of aggregated experience from all members.

We evaluate GEA on challenging coding benchmarks, achieving success rates of 71.0\% on SWE-bench Verified and 88.3\% on Polyglot, significantly outperforming state-of-the-art open-ended self-evolving methods (56.7\% and 68.3\%, respectively). Analysis reveals that GEA more effectively consolidates the diversity generated during open-ended exploration, yielding sustained progress and stronger performance given the same number of evolved agents. By leveraging experience from better-performing agents, GEA also exhibits stronger robustness to framework-level perturbations. Furthermore, its improvements stem from workflow and tool enhancements rather than model-specific optimizations, thus transferring consistently across GPT- and Claude-series models.

Additionally, by leveraging meta-learning for self-improvement in open-ended exploration, without any human intervention, GEA achieves performance comparable to or even surpassing human-designed state-of-the-art frameworks on both benchmarks (71.0\% vs. 71.8\% on SWE-bench Verified, 88.3\% vs. 52.0\% on Polyglot).

In summary, we propose \textbf{Group-Evolving Agents}, a new paradigm for open-ended self-improvement that:

\begin{enumerate}
    \item Overcomes the limitation of inefficient utilization of exploratory diversity caused by branch isolation in existing tree-structured evolution, by enabling explicit experience sharing and reuse within the group during evolution.
    \item More effectively consolidates and reuses experience and evolutionary diversity from other agents, achieving significant performance gains and stronger robustness over state-of-the-art open-ended self-evolving methods, with improvements that transfer consistently across different coding models.
    \item Matches or surpasses human-designed state-of-the-art frameworks through meta-learning-based self-improvement without human intervention.
\end{enumerate}
\section{Related Work}

Recent years have witnessed growing interest in how AI systems can continuously improve themselves without human intervention~\citep{fang2025comprehensive,gao2025survey,wang2025evoagentx}. Most existing self-improving approaches mainly focus on continuous, iterative refinement of the given agent system~\citep{shinn2023reflexion,songmind,gao2025survey,zelikman2024self,singhbeyond}, typically evolving toward a specific optimization objective and following a linear, chain-based evolutionary structure~\citep{zhang2025population,wang2025evoagentx,robeyns2025self} . Such systems achieve self-improvement through mechanisms such as self-play against historical versions or self-generated verification~\citep{setlur2024rewarding,wang2024math,huang2025r,wang2025socratic,wei2025toward}, supervised fine-tuning~\citep{zelikman2022star,gou2023tora,ni2024next} or reinforcement learning on selectively filtered feedback~\citep{xin2024deepseek,zhao2025absolute,yuan2024self} , and reflection-based methods~\citep{wang2025mobile,yin2025godel,shinn2023reflexion} or in-context learning~\citep{hua2024trustagent,sun2023adaplanner}. While this goal-oriented, chain-based evolutionary paradigm enables autonomous improvement along a particular direction, it inherently limits the ability of self-evolving systems to explore diverse evolutionary directions in open-ended solution spaces.

A line of work has pointed out that one of the key challenges in enabling unbounded improvement and innovation lies in developing open-ended AI systems that can continuously produce both novel and learnable artifacts~\citep{stanley2017open,zhang2025darwin,stanley2015greatness,shapere1964structure}. Building on this insight, open-endedness has been characterized as the capability of systems to continuously generate artifacts that are novel, interesting, and learnable from a human perspective~\citep{zhang2025darwin,clune2019ai,hughes2024open,jiang2023general,zhangomni,faldoromni}.

Motivated by the potential of enabling unbounded evolution through open-ended exploration in self-evolving agents, more recent studies adopt lineage-based, tree-structured evolutionary strategies~\citep{antoniadesswe} inspired by biological inheritance and mutation~\citep{zhang2025darwin,novikov2025alphaevolve,fernando2024promptbreeder,huang2023large}. In these frameworks, individual parent agents are selected at each iteration to independently produce offspring, enabling various branching exploration across multiple evolutionary directions and helping avoid local optima. However, the strict isolation between evolutionary branches prevents effective information and experience sharing and reuse across lineages. As a result, many promising directions discovered early in evolution persist only as temporary diversity and fail to contribute to long-term cumulative progress.
To overcome this limitation, we introduce a group-centric evolutionary paradigm, \textbf{Group-Evolving Agents (GEA)}, which explicitly enables intra-group experience sharing and reuse throughout the evolutionary process. By consolidating complementary discoveries across agents, GEA more effectively leverages the diversity generated by open-ended exploration to support sustained cumulative progress.

\section{Method}
We propose \textbf{G}roup-\textbf{E}volving \textbf{A}gents,  a framework for open-ended evolution that treats a \emph{group of agents} as the fundamental unit of evolution. GEA maintains an archive that stores all discovered agents throughout the evolutionary process. As shown in Figure~\ref{fig:pipeline-overall}, at each iteration, GEA proceeds in two core stages:

\textbf{(1) Parent Group Selection} (\S\ref{sec:parent-selection}): GEA first selects $K$ parent agents from the archive using a Performance--Novelty selection strategy~\citep{mouret2015illuminating,pugh2016quality,chatzilygeroudis2021quality} that balances immediate task-solving competence with long-term evolutionary diversity and potential.

\textbf{(2) Open-ended Group Evolution} (\S\ref{sec:group-evolution}): The selected agents form a parent group that jointly produces an offspring group of the same size through explicit experience sharing and reuse across parent agents.

We detail the method below.

\begin{algorithm}[htbp]
\caption{Parent Group Selection with KNN Novelty}
\label{alg:parent-group-selection}
\begin{algorithmic}[1]

\STATE \textbf{Input:} Archive of agents $\mathcal{A}$; agent-representation vectors $\{z_i \in \{0,1\}^D\}_{i\in\mathcal{A}}$;
performance scores $\{\alpha_i\}_{i\in\mathcal{A}}$;
parent group size $K$;
KNN size $M$.
\STATE \textbf{Output:} Parent agent group $\mathcal{G}$ with $|\mathcal{G}| = K$.

\vspace{0.5em}

\STATE \textbf{/* Compute novelty for each agent */}
\FOR{$i \in \mathcal{A}$}
    \STATE Initialize empty list $\mathcal{D}_i$
    \FOR{$j \in \mathcal{A},\ j \neq i$}
        \STATE $d_{ij} \leftarrow 1 - \dfrac{z_i^\top z_j}{\|z_i\|_2\,\|z_j\|_2 + \varepsilon}$
        \STATE Append $d_{ij}$ to $\mathcal{D}_i$
    \ENDFOR
    \STATE Let $\mathcal{N}_M(i)$ be the indices of the $M$ smallest values in $\mathcal{D}_i$
    \STATE $\text{nov}(i) \leftarrow \dfrac{1}{M} \sum_{j \in \mathcal{N}_M(i)} d_{ij}$
\ENDFOR

\vspace{0.5em}

\STATE \textbf{/* Rank agents by Performance--Novelty score */}
\FOR{$i \in \mathcal{A}$}
    \STATE $\text{score}(i) \leftarrow \alpha_i \cdot \sqrt{\text{nov}(i)}$
\ENDFOR

\STATE $\mathcal{G} \leftarrow$ the top-$K$ agents in $\mathcal{A}$ ranked by $\text{score}(\cdot)$
\STATE \textbf{return} $\mathcal{G}$

\end{algorithmic}
\end{algorithm}

\subsection{Parent Group Selection}
\label{sec:parent-selection}
Inspired by~\citet{mouret2015illuminating,pugh2016quality,chatzilygeroudis2021quality}, parent group selection in GEA balances two key principles: \emph{performance} and \emph{novelty}. 
We prioritize agents with strong task performance, as performance reflects an agent's immediate competence and its likelihood of producing effective offspring, since evolution in GEA proceeds through iterative modifications of the agent's implementation, which itself constitutes a form of solving coding problems. At the same time, we also encourage exploration beyond currently well-optimized regions of the search space, as agents that exhibit novel evolutionary directions may contribute to long-term cumulative progress even when their current performance is not optimal.

We represent each agent $i$ using a task-success vector $z_i \in \{0,1\}^D$, where each dimension indicates whether the agent successfully solves a corresponding probe task. Similar binary task–response representations of this form have been widely used to characterize an agent’s coding capabilities and to better understand how these capabilities are distributed across various tasks~\cite{polo2024tinybenchmarks,rodriguez2021evaluation}. Using this representation, we measure the dissimilarity between two agents via cosine distance:
\begin{equation}
d(i,j) = 1 - \frac{z_i^\top z_j}{\lVert z_i \rVert_2 \, \lVert z_j \rVert_2+ \varepsilon}.
\end{equation}

We define the novelty of agent $i$ as the average cosine distance to its $M$ most similar neighbors:
\begin{equation}
\mathrm{nov}(i) = \frac{1}{M} \sum_{j \in \mathcal{N}_M(i)} d(i,j),
\end{equation}
where $\mathcal{N}_M(i)$ denotes the set of $M$ agents with the smallest cosine distance to agent $i$.

To construct the parent group, we rank agents according to a combined score
\begin{equation}
\mathrm{score}(i) = \alpha_i \cdot \sqrt{\mathrm{nov}(i)},
\end{equation}
where $\alpha_i$ denotes the performance of agent $i$ on downstream coding tasks, and $\sqrt{\mathrm{nov}(i)}$ moderates the influence of novelty.
Finally, we select the top-$K$ agents according to this score to form the parent group. Performance serves as the primary selection criterion, while novelty is incorporated as a mild bias without dominating performance, enabling a balanced trade-off between exploitation and exploration. The full procedure is summarized in Algorithm~\ref{alg:parent-group-selection}.

% \noindent
% \begin{minipage}[t]{0.48\textwidth}
% % 左边继续写你的文字（或者把这一段文字放到这里）
% Given a selected parent group $\mathcal{G}=\{a_1,\dots,a_K\}$, ...
% \end{minipage}
% \hfill
% \begin{minipage}[t]{0.48\textwidth}
% \centering
% \includegraphics[width=\linewidth]{graph/pipeline2.pdf}
% \captionof{figure}{\textbf{Detailed illustration of group-level evolution in GEA.} Aggregated evolutionary traces from the parent group are shared across all agents to generate evolution directives and framework-level patches.}
% \label{fig:pipeline-detailed}
% \end{minipage}

\begin{algorithm}[htbp]
\caption{Open-Ended Group-Evolving}
\label{alg:group-evolution}
\begin{algorithmic}[1]

\STATE \textbf{Input:} Parent group $\mathcal{G}=\{a_1,\dots,a_K\}$; archive $\mathcal{A}$; coding benchmark $\mathcal{T}$
\STATE \textbf{Output:} Offspring group $\mathcal{G}'$ with $|\mathcal{G}'|=K$; updated archive $\mathcal{A}$

\vspace{0.5em}
\STATE Initialize offspring group $\mathcal{G}' \leftarrow \emptyset$

\vspace{0.5em}
\FOR{$a_i \in \mathcal{G}$}
    \STATE \textbf{/* Collect evolutionary traces */}
    \STATE $\mathcal{P}_i^{\mathrm{Applied}} \leftarrow$ \textsc{GetAppliedPatches}$(a_i)$
    \STATE $t_i^{\mathrm{fail}} \leftarrow$ \textsc{SampleUnsolvedTask}$(a_i)$
    \STATE $\mathcal{P}_i^{\mathrm{pred}} \leftarrow$ \textsc{GetPredictedTaskPatch}$(a_i, t_i^{\mathrm{fail}})$
    \STATE $\mathcal{L}_i \leftarrow$ \textsc{GetExecutionLogs}$(a_i, t_i^{\mathrm{fail}})$
    \STATE $\mathcal{O}_i \leftarrow$ \textsc{GetOutcomeLog}$(a_i, t_i^{\mathrm{fail}})$
    \STATE $T_i \leftarrow \{\mathcal{P}_i^{\mathrm{Applied}}, \mathcal{P}_i^{\mathrm{pred}}, \mathcal{L}_i, \mathcal{O}_i\}$
\ENDFOR

\vspace{0.5em}
\STATE \textbf{/* Aggregate and share group-level experience */}
\STATE $S \leftarrow \bigcup_{a_j \in \mathcal{G}} T_j$

\vspace{0.5em}
\FOR{$a_i \in \mathcal{G}$}
    \STATE \textbf{/* Reflection: analyze shared experience */}
    \STATE $\Delta_i \leftarrow$ \textsc{Reflect}$(a_i;\, S)$
    \hfill // evolution directives

    \vspace{0.25em}
    \STATE \textbf{/* Evolution: generate framework-level patches */}
    \STATE $\pi_i' \leftarrow$ \textsc{Evolve}$(a_i;\, \Delta_i)$

    \vspace{0.25em}
    \STATE \textbf{/* Acting: evaluate updated agent */}
    \STATE $a_i' \leftarrow$ \textsc{ApplyPatch}$(a_i, \pi_i')$
    \STATE \textsc{ActAndEvaluate}$(a_i';\, \mathcal{T})$

    \vspace{0.25em}
    \STATE $\mathcal{G}' \leftarrow \mathcal{G}' \cup \{a_i'\}$

    \STATE \textbf{/* Archive update */}
    \IF{\textsc{Compiles}$(a_i')$ \AND \textsc{BasicCodingFunc}$(a_i')$}
        \STATE $\mathcal{A} \leftarrow \mathcal{A} \cup \{a_i'\}$
    \ENDIF
\ENDFOR

\vspace{0.5em}
\STATE \textbf{return} $\mathcal{G}', \mathcal{A}$

\end{algorithmic}
\end{algorithm}

\subsection{Open-Ended Group Evolution}
\label{sec:group-evolution}

Unlike conventional approaches where parent agents evolve independently without information and experience exchange, GEA explicitly enables experience sharing and reuse among agents during evolution. This group-level experience sharing allows agents to integrate complementary evolutionary directions explored by different agents while maintaining open-ended exploration. Diversity generated during exploration is thus transformed from transient variations into long-term useful experience, effectively contributing to sustained evolutionary progress.

% \begin{wrapfigure}{r}{0.48\linewidth}
%   \vspace{-6pt} % 可微调
%   \centering
%   \includegraphics[width=\linewidth]{graph/pipeline2.pdf}
%   \vspace{-8pt}
%   \captionof{figure}{\textbf{Detailed illustration of group-level evolution in GEA.} Aggregated evolutionary traces from the parent group are shared across all agents to generate evolution directives and framework-level patches.}
%   \vspace{-10pt}
% \end{wrapfigure}

\begin{wrapfigure}{14}{0.48\linewidth}
  \vspace{-6pt} % 可微调
  \centering
  \includegraphics[width=\linewidth]{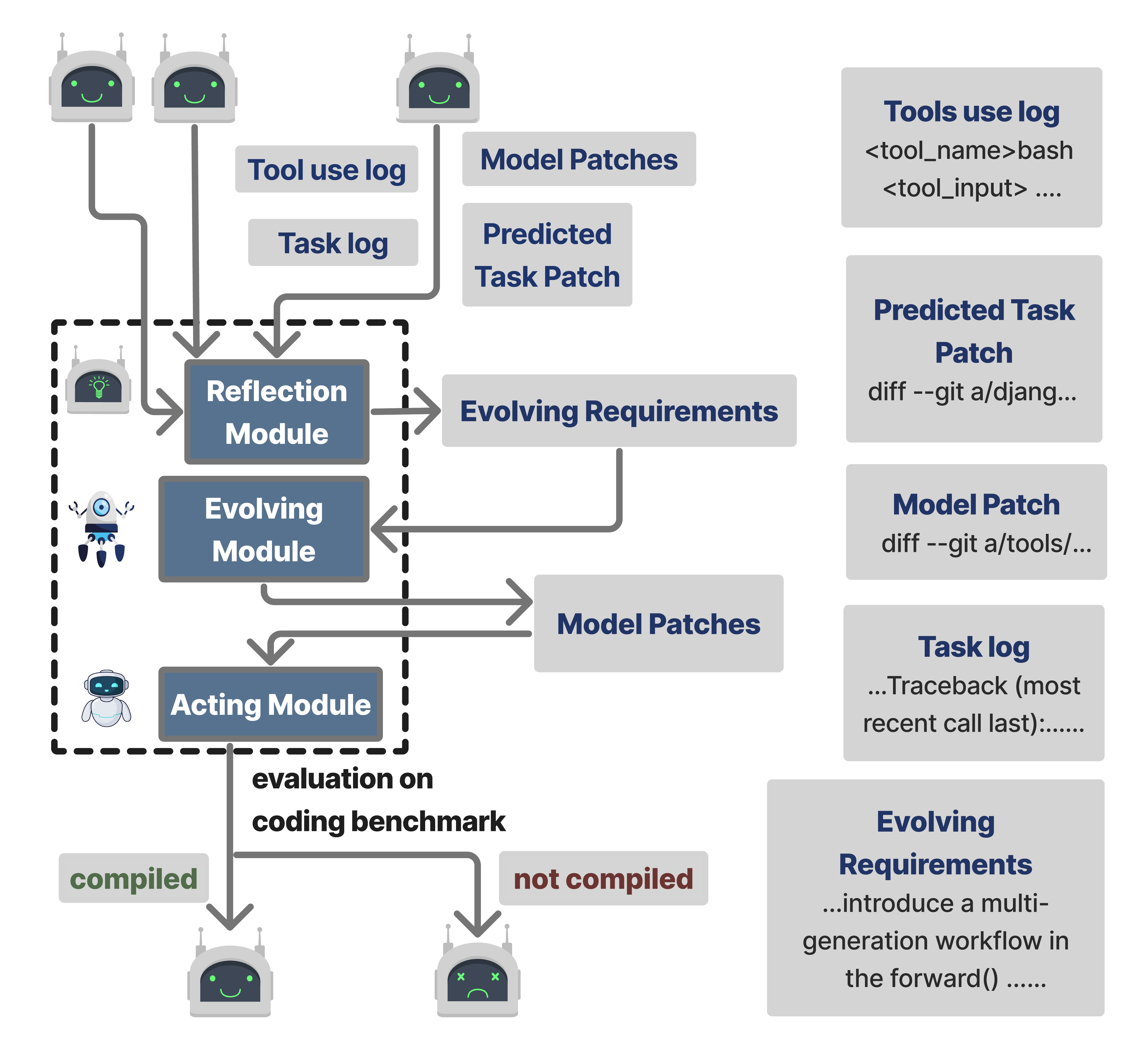}
  \vspace{-8pt}
  \captionof{figure}{\textbf{Detailed illustration of group-level evolution in GEA.} Aggregated evolutionary traces from the parent group are shared across all agents to generate evolution directives and framework-level patches.}
  \label{fig:pipeline-detailed}
  \vspace{-10pt}
\end{wrapfigure}

Given a selected parent group $\mathcal{G} = \{a_1, a_2, \ldots, a_K\}$, GEA generates a new group $\mathcal{G}'$ of the same size, where each agent evolves by leveraging both its own evolutionary history and experience aggregated from other members of the parent group, as demonstrated in Figure~\ref{fig:pipeline-detailed}.

For each agent $a_i \in \mathcal{G}$, we collect a set of evolutionary traces consisting of:
\begin{enumerate}
    \item the code modification patches applied to the agent's framework;
    \item a predicted task patch generated by $a_i$ for a randomly sampled unsolved task during evaluation;
    \item the corresponding task execution logs, including the complete tool invocation history and execution workflow;
    \item the evaluation outcome of the same task, which exposes failure modes and potential directions for framework-level improvement.
\end{enumerate}

The aggregated traces from all agents in the parent group are provided as shared input to every agent. Each agent evolves from this shared pool of group-level experience while diverging through complementary adaptations to its own codebase, enabling the group to explore diverse evolutionary directions while leveraging experiences from one another.

For each agent $a_i$, the shared group-level experience is fed into its reflection module, which analyzes these traces and produces evolution directives targeting the agent's workflow, tool usage, or prompting strategies. These directives are then passed to the evolution module to generate framework-level patches. Finally, the updated agent is evaluated on downstream programming tasks via the action module. Agents that compile successfully and exhibit basic coding functionality are retained and added to the archive for future evolution.

Applying this process to each agent in the parent group yields an offspring group of size $K$. This group-level evolution iterates in an open-ended manner, as summarized in Algorithm~\ref{alg:group-evolution}.
\section{Experiments}

\subsection{Benchmarks}

Following the evaluation protocol established by the state-of-the-art open-ended self-evolving system, Darwin Gödel Machine (DGM)~\citep{zhang2025darwin}, we evaluate GEA on two structurally distinct benchmarks to assess its coding capabilities in both repository-level software engineering and multi-language code synthesis settings. To mitigate the substantial cost of evaluating every evolved agent on the full benchmarks, we adopt a staged evaluation strategy where agents must pass smaller subsets before advancing.

\paragraph{SWE-bench.}
We evaluate on \textbf{SWE-bench Verified}~\citep{swe-bench-verified}, a curated, human-validated subset of SWE-bench in which every task is confirmed to be solvable. We employ a three-stage evaluation process. First, agents undergo a sanity check on a small set of 10 tasks to discard those with framework-level failures  (i.e., cannot compile or solve 0 out of 10 tasks). Agents that demonstrate basic coding functionality are then evaluated on the 50-task \textit{Verified-Mini} set~\citep{swebench-verified-mini}, which is designed to preserve a similar difficulty and pass-rate distribution as the full benchmark with fewer samples. Finally, the top-2 performing agents from this stage are evaluated on the full SWE-bench Verified dataset.

\paragraph{Polyglot.}
Polyglot~\citep{aider-github,aider-polyglot} assesses algorithmic code generation across diverse languages (C++, Rust, Java, etc.) and serves as an out-of-domain generalization test, since it is rarely used for training or fine-tuning coding models~\citep{zhang2025darwin}. We report pass@1 performance. Unlike SWE-bench Verified~\citep{swe-bench-verified}, group-level evolution is conducted exclusively on a 10-task small set. Agents achieving a success rate above 40\% are subsequently evaluated on a separate, unseen 50-task medium set, ensuring that the larger set remains a strict out-of-distribution test for generalization.

\subsection{Experimental Settings}
\label{sec:experimental_settings}

For both SWE-bench Verified and Polyglot, we set the group size to $K=2$
and use $M=4$ nearest neighbors in the KNN-based parent group selection.
At each iteration, a parent group of size 2 produces an offspring group of the same size.

For SWE-bench Verified, we run group-evolution for 30 iterations.
Due to computational budget constraints, during the first 20 iterations,
the evolving and acting modules are powered by Claude Haiku~4.5~\citep{anthropic-claude-haiku-4-5},
while the final 10 iterations use Claude Sonnet~4.5~\citep{anthropic-claude-sonnet-4-5}.
The reflection module is consistently powered by GPT-o1 across all iterations~\citep{jaech2024openai}. For Polyglot, we run group-evolution for 20 iterations.
The first 10 iterations use Claude Haiku 4.5~\citep{anthropic-claude-haiku-4-5} for evolving and acting module, and the remaining 10 iterations use Claude Sonnet 4.5~\citep{anthropic-claude-sonnet-4-5}.
The reflection module is again consistently powered by GPT-o1~\citep{jaech2024openai}.

\subsection{Baselines}
\label{sec:baselines_settings}

We use state-of-the-art open-ended self-evolving agents as our primary baseline for comprehensive comparison, to systematically examine how experience sharing affects diversity utilization and performance improvement during open-ended evolution, as well as robustness to framework-level perturbations. Additionally, we compare the final performance of GEA, which uses a meta-learning approach without any human intervention, against state-of-the-art human-designed coding agents.

\paragraph{Open-Ended Self-Evolving Baseline.}
We compare against the current state-of-the-art open-ended self-evolving framework, implemented following DGM~\citep{zhang2025darwin}. Unlike GEA, this baseline enforces a strict tree-structured evolution, where only a single agent $a_i$ is selected as the parent and independently evolves into one child agent at each iteration. Evolutionary experience is not shared across different evolutionary branches. Specifically, the reflection module of agent $a_i$ receives only its own evolutionary traces, including:
\begin{enumerate}[label=(\roman*), leftmargin=*, align=left]
    \item code modification patches applied to the agent's framework;
    \item a predicted task patch for a randomly sampled unsolved task;
    \item the corresponding execution log, including tool invocation history and workflow; and
    \item the evaluation outcome, exposing failure modes and improvement directions.
\end{enumerate}
This design prevents experience reuse across evolutionary branches, resulting in a strictly individual-centric evolutionary process.

For SWE-bench Verified~\citep{swe-bench-verified}, we run this baseline for 60 iterations in total: the evolution and coding modules are powered by Claude Haiku~4.5~\citep{anthropic-claude-haiku-4-5} for the first 40 iterations and Claude Sonnet~4.5~\citep{anthropic-claude-sonnet-4-5} for the final 20 iterations. For Polyglot, we run the baseline for 40 iterations: Claude Haiku~4.5~\citep{anthropic-claude-haiku-4-5} for the first 20 iterations and Claude Sonnet~4.5~\citep{anthropic-claude-sonnet-4-5} for the remaining 20. In all baseline experiments, the reflection module is consistently powered by GPT-o1~\citep{jaech2024openai}. To ensure fair comparison, we intentionally run the baseline for twice as many iterations as GEA so that the total number of evolved agents is comparable across methods, ensuring all comparisons are conducted under matched model schedules.

\paragraph{Human-Designed Frameworks.}
We additionally compare against state-of-the-art human-designed frameworks on both benchmarks. The top-performing, open-scaffold, checked entry on SWE-bench Verified is OpenHands + GPT-5~\citep{wangopenhands,singh2025openai}, achieving 71.8\%, where ``checked'' indicates that the SWE-bench team successfully reproduced the reported patch generations~\citep{jimenez2024swe,zhang2025darwin}.  For Polyglot, which was originally used to evaluate Aider~\citep{aider-polyglot,aider-github} by its developers, we compare against Aider, a widely adopted coding agent under continuous development and testing by human developers. The state-of-the-art performance is Aider + GPT-5 (high)~\citep{aider-polyglot,singh2025openai}, achieving a 52.0\% pass@1 success rate.
\section{Results and Analysis}

\subsection{Main Results}

\begin{figure*}[t]
\centering

\begin{subfigure}[t]{0.48\textwidth}
  \vspace{0pt}
  \centering
  \includegraphics[width=\linewidth]{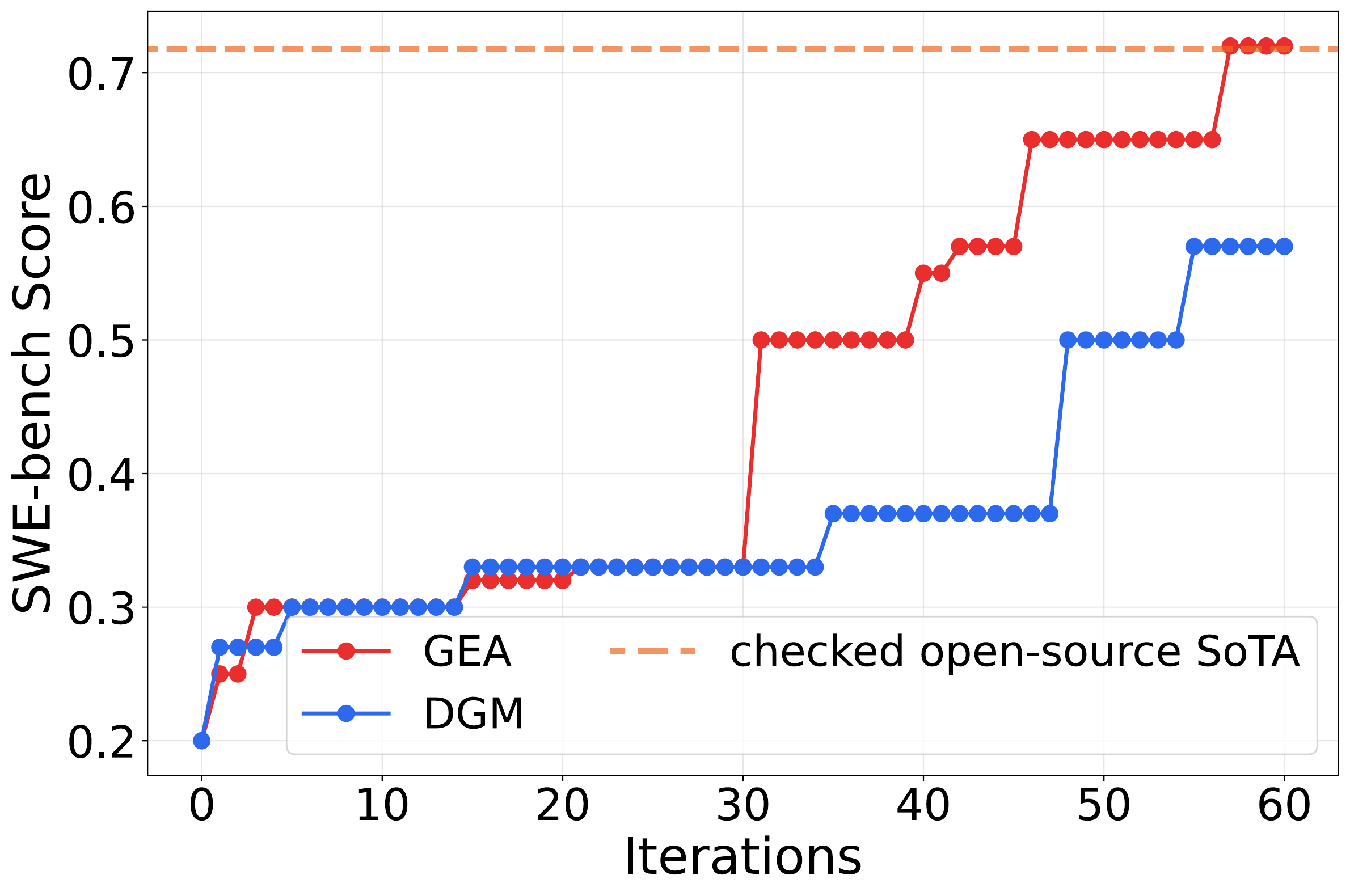}
  \caption{GEA vs.\ DGM (Self-evolving Baseline) on SWE-bench}
  \label{fig:result-a}
\end{subfigure}\hfill
\begin{subfigure}[t]{0.48\textwidth}
  \vspace{0pt}
  \centering
  \includegraphics[width=\linewidth]{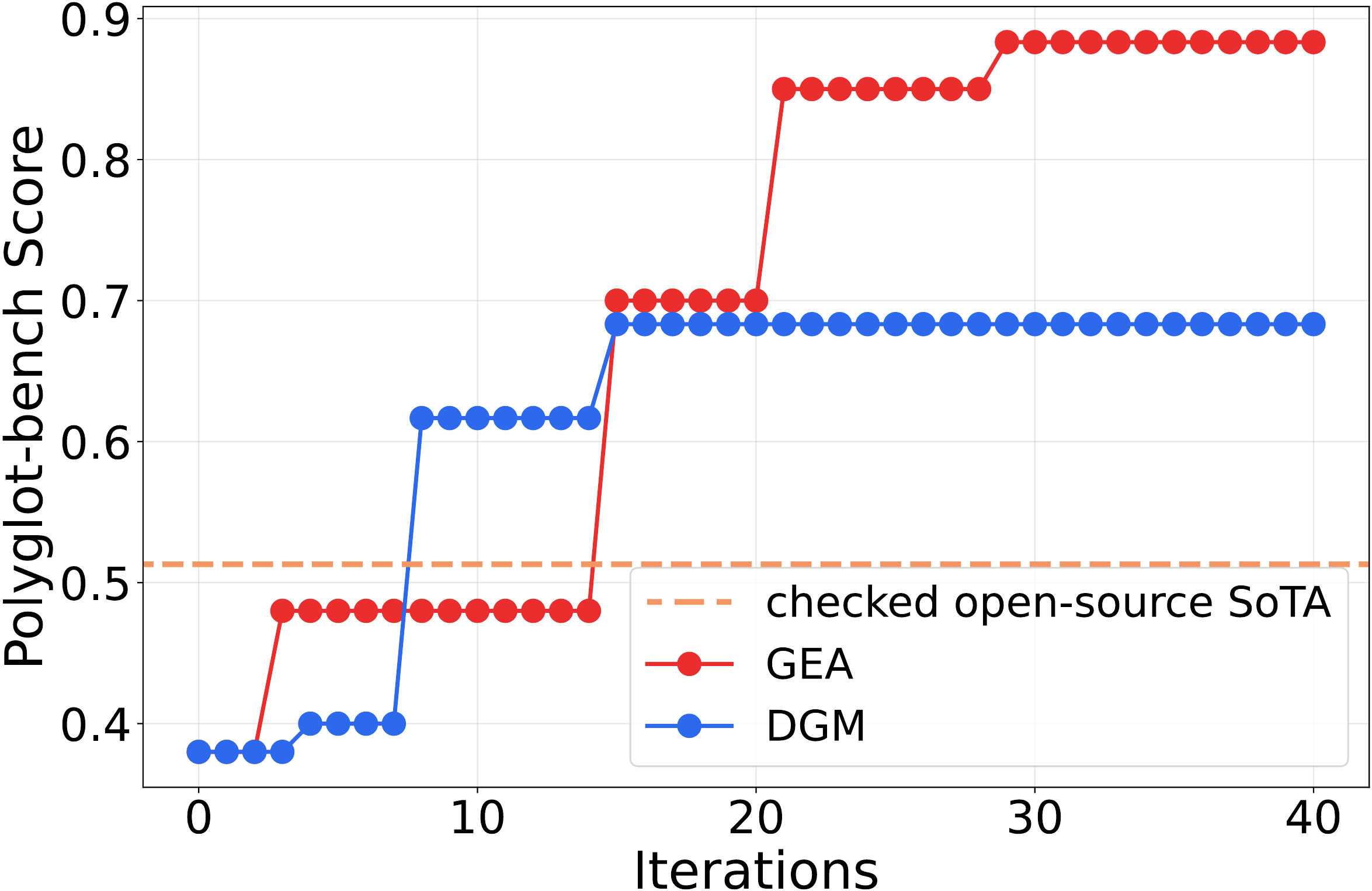}
  \caption{GEA vs.\ DGM (Self-evolving Baseline) on Polyglot}
  \label{fig:result-b}
\end{subfigure}

\caption{
\textbf{Performance comparison between GEA and DGM (self-evolving baseline) on two coding benchmarks.} Under the same number of evolved agents, GEA exhibits substantially larger performance gains than DGM on both SWE-bench and Polyglot, demonstrating the improved efficiency of group-level evolution.
}
\label{fig:result}
\end{figure*}

\paragraph{GEA vs.\ State-of-the-Art Open-Ended Self-Evolving Systems.}
As shown in Figure~\ref{fig:result}, GEA demonstrates substantial performance improvements over the DGM (self-evolving baseline) on both SWE-bench Verified and Polyglot. On SWE-bench Verified, GEA improves performance from 20.0\% to 71.0\%, while under the same number of evolved agents, the DGM baseline achieves only 56.7\%. On Polyglot, GEA boosts performance from 38.2\% to 88.3\%, significantly outperforming the DGM baseline (68.3\%). Notably, GEA exhibits faster and more pronounced improvement in the mid-to-late stages of evolution compared to DGM~\citep{zhang2025darwin}, one potential reason is that the archive has accumulated sufficient diverse evolutionary directions by this point, that can be progressively consolidated and reused, leading to more rapid and pronounced performance gain, aligning with Figure~\ref{fig:evolution-analysis}. We discuss this phenomenon in detail in Section~\ref{sec:evolution}.

\paragraph{GEA vs.\ State-of-the-Art Human-Designed Agents.}
GEA achieves performance comparable to or exceeding state-of-the-art human-designed agents on both benchmarks: 71.0\% vs.\ 71.8\% on SWE-bench Verified, and 88.3\% vs.\ 52.0\% on Polyglot. Using meta-learning without any human intervention, GEA automatically evolves agent frameworks that match or surpass carefully engineered human designs~\citep{aider-github,singh2025openai,wangopenhands}, demonstrating the potential and viability of fully autonomous agent improvement.

\paragraph{Analysis of evolutionary patterns on two benchmarks.} GEA achieves state-of-the-art open-ended self-improvement on both benchmarks, with particularly strong performance on Polyglot~\citep{aider-github,aider-polyglot}. We further analyze and observe that the two benchmarks differ in task complexity: Polyglot~\citep{aider-github,aider-polyglot} primarily requires agents to modify a single file from scratch to resolve tasks, involving lower editing complexity without multi-file coordination. In contrast, SWE-bench Verified~\citep{swe-bench-verified,jimenez2024swe} typically requires coordinated modifications across multiple files, demanding that agents understand inter-file dependencies and locate relevant files for coordinated edits.

This difference in complexity leads to distinct evolutionary patterns. On Polyglot~\citep{aider-github,aider-polyglot}, meta-learning produces larger, more concentrated patches: each iteration yields substantial performance gains, reaching 88.3\% in just 4 iterations with a total of 8,677 lines of code added. On SWE-bench Verified~\citep{swe-bench-verified,jimenez2024swe}, the evolved patches are smaller and more distributed, requiring 8 iterations to reach 71.0\% with 9,663 lines of code added. These observations suggest that GEA adapts its evolutionary behavior to varying task complexity, demonstrating the flexibility and generality of group-level meta-learning across different problem settings.

\subsection{Evolution Analysis}
\label{sec:evolution}

\begin{figure}[t]
\centering

% ===== 左：Figure =====
\begin{minipage}[t]{0.48\columnwidth}
  \vspace{0pt} % 关键：顶端对齐
  \centering
  \includegraphics[width=\linewidth]{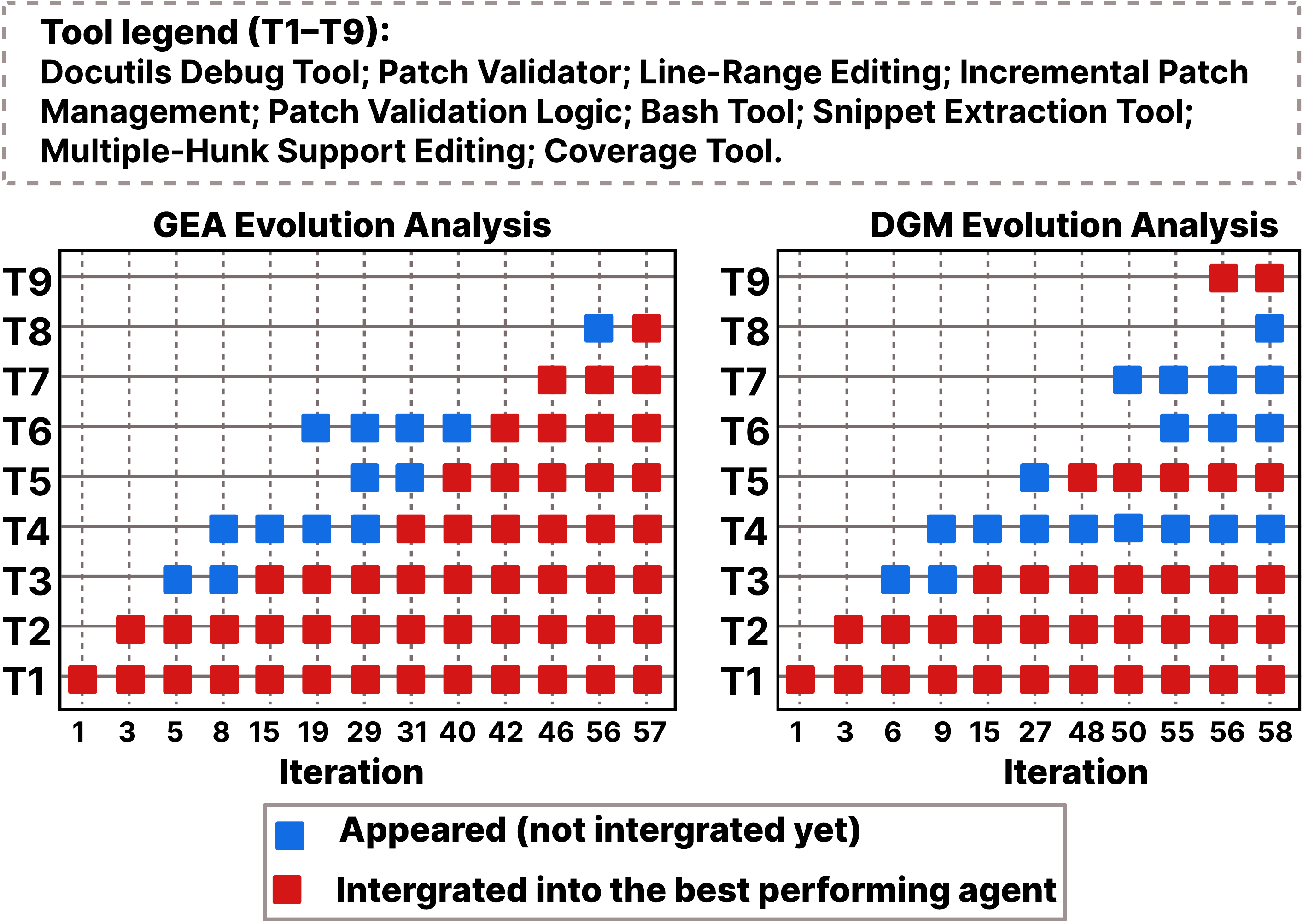}

  \captionof{figure}{\textbf{Evolution analysis of tool discovery and integration over iterations.}
  Each row (T1--T9) corresponds to a key tool-level functionality.
  Blue markers indicate tools that have been discovered but not yet integrated into the current best agent,
  while red markers indicate tools integrated into the best-performing agent.}
  \label{fig:evolution-analysis}
\end{minipage}
\hfill
% ===== 右：Table =====
\begin{minipage}[t]{0.48\columnwidth}
  \vspace{0pt} % 关键：顶端对齐
  \centering
  \small
  \setlength{\tabcolsep}{4pt}
  \renewcommand{\arraystretch}{1.15}
  \begin{tabular}{l cc | cc}
    \toprule
    & \multicolumn{2}{c}{\textbf{Worst-Case Success} $\uparrow$}
    & \multicolumn{2}{c}{\textbf{Ancestor Count} $\uparrow$} \\
    \cmidrule(lr){2-3} \cmidrule(lr){4-5}
    \textbf{Rank} & \textbf{DGM} & \textbf{GEA} & \textbf{DGM} & \textbf{GEA} \\
    \midrule
    Top-1 & 56.7\% & \textbf{71.0\%} & 9 \scriptsize{(15.0\%)} & \textbf{17} \scriptsize{(28.3\%)} \\
    Top-3 & 48.3\% & \textbf{63.3\%} & 13 \scriptsize{(21.7\%)} & \textbf{17} \scriptsize{(28.3\%)} \\
    Top-5 & 45.0\% & \textbf{58.3\%} & 16 \scriptsize{(26.7\%)} & \textbf{18} \scriptsize{(30.0\%)} \\
    \bottomrule
  \end{tabular}

  \captionof{table}{\textbf{Comparison of performance (Success Rate) and ancestor integration across the Top-k agents on SWE-bench Verified}. Performance is reported as the \emph{worst-case (minimum) success rate} among the top-$k$ agents. \emph{Ancestor Count} denotes the count of unique historical agents integrated into the solution. Notably, the worst-case performance of GEA's top-5 agents (58.3\%) exceeds the single best agent from DGM (56.7\%).}
  \label{tab:ancestor-integration}
\end{minipage}

\end{figure}

Overall, our analysis shows that GEA can efficiently consolidate tool-level innovations discovered across the agents, rather than letting them remain isolated in separate evolutionary branches. Figure~\ref{fig:evolution-analysis} summarizes nine key tool-level modifications on agents' framework that drove improvements. GEA integrated eight of these functionalities into its best agent, whereas the best DGM agent integrated only five. Crucially, the four tools missing from the DGM agent \textit{were} explored in isolated branches (e.g., \textbf{T4} at iteration 9) but failed to propagate due to lineage isolation. In contrast, GEA systematically consolidated these dispersed capabilities; five of its integrated tools originated from different parent agents, confirming that explicit experience sharing prevents beneficial innovations from dying out.

% \begin{wraptable}{r}{0.48\columnwidth}
%   \vspace{-6pt}
%   \centering
%   \small
%   \setlength{\tabcolsep}{4pt}
%   \renewcommand{\arraystretch}{1.15}
%   \begin{tabular}{l cc | cc}
%     \toprule
%     & \multicolumn{2}{c}{\textbf{Worst-Case Success} $\uparrow$} & \multicolumn{2}{c}{\textbf{Ancestor Count} $\uparrow$} \\
%     \cmidrule(lr){2-3} \cmidrule(lr){4-5}
%     \textbf{Rank} & \textbf{DGM} & \textbf{GEA} & \textbf{DGM} & \textbf{GEA} \\
%     \midrule
%     Top-1 & 56.7\% & \textbf{71.0\%} & 9 \scriptsize{(15.0\%)} & \textbf{17} \scriptsize{(28.3\%)} \\
%     Top-3 & 48.3\% & \textbf{63.3\%} & 13 \scriptsize{(21.7\%)} & \textbf{17} \scriptsize{(28.3\%)} \\
%     Top-5 & 45.0\% & \textbf{58.3\%} & 16 \scriptsize{(26.7\%)} & \textbf{18} \scriptsize{(30.0\%)} \\
% \bottomrule
%   \end{tabular}
%   \vspace{-4pt}
%   \caption{\textbf{Comparison of performance (Success Rate) and ancestor integration across the Top-k agents on SWE-bench Verified}. Performance is reported as the \emph{worst-case (minimum) success rate} among the top-$k$ agents. \emph{Ancestor Count} denotes the count of unique historical agents integrated into the solution. Notably, the worst-case performance of GEA's top-5 agents (58.3\%) exceeds the single best agent from DGM (56.7\%).}
%   \label{tab:ancestor-integration}
%   \vspace{-10pt}
% \end{wraptable}

To quantify this consolidation, we track the number of unique \textit{ancestor agents} contributing to the final solutions (Table~\ref{tab:ancestor-integration}). The best GEA agent integrates experiences from 17 unique ancestors (28.3\% of the population)—nearly double that of the best DGM agent (9 ancestors). 

This broader integration correlates directly with population-wide quality. As shown in Table~\ref{tab:ancestor-integration}, we report the \emph{worst-case} performance among the top-$k$ agents. Notably, the \textbf{worst} of GEA's top-5 agents achieves 58.3\%, which strictly outperforms the \textbf{single best} agent produced by DGM (56.7\%). This confirms that GEA does not merely produce outliers, but systematically elevates the entire population by effectively consolidating complementary and diverse evolutionary paths. More broadly, this efficient experience consolidation suggests that GEA may exhibit stronger evolutionary capabilities in environments with greater diversity. Since open-ended evolution naturally leads to a monotonically growing diversity in the archive, GEA not only achieves stronger performance as evolution progresses but also gains enhanced self-improvement capabilities, consistent with the results shown in Figure~\ref{fig:result}.

% \begin{table}[h]
% \centering
% \begin{small}
% \begin{tabular}{l cc | cc}
% \toprule
% & \multicolumn{2}{c}{\textbf{Worst-Case Success} $\uparrow$} & \multicolumn{2}{c}{\textbf{Ancestor Count} $\uparrow$} \\
% \cmidrule(lr){2-3} \cmidrule(lr){4-5}
% \textbf{Rank} & \textbf{DGM} & \textbf{GEA} & \textbf{DGM} & \textbf{GEA} \\
% \midrule
% Top-1 & 56.7\% & \textbf{71.0\%} & 9 \scriptsize{(15.0\%)} & \textbf{17} \scriptsize{(28.3\%)} \\
% Top-3 & 48.3\% & \textbf{63.3\%} & 13 \scriptsize{(21.7\%)} & \textbf{17} \scriptsize{(28.3\%)} \\
% Top-5 & 45.0\% & \textbf{58.3\%} & 16 \scriptsize{(26.7\%)} & \textbf{18} \scriptsize{(30.0\%)} \\
% \bottomrule
% \end{tabular}
% \vspace{4pt}
% \caption{
% \textbf{Comparison of performance (Success Rate) and ancestor integration across the Top-k agents on SWE-bench Verified}. Performance is reported as the \emph{worst-case (minimum) success rate} among the top-$k$ agents. \emph{Ancestor Count} denotes the count of unique historical agents integrated into the solution. Notably, the worst-case performance of GEA's top-5 agents (58.3\%) exceeds the single best agent from DGM (56.7\%).
% }
% \label{tab:ancestor-integration}
% \end{small}
% \end{table}

\subsection{Transferability}

 We evaluate the generalization and model transferability of the best-discovered agent produced by GEA.
Specifically, we replace the coding model used in the acting module with different GPT-series and Claude-series models during benchmark evaluation.
We then compare the performance of the initial(iteration--0) agent and the GEA best-discovered agent under each coding model.

\begin{wrapfigure}{r}{0.48\columnwidth}
  \vspace{-6pt} % 可微调
  \centering
  \includegraphics[width=\linewidth]{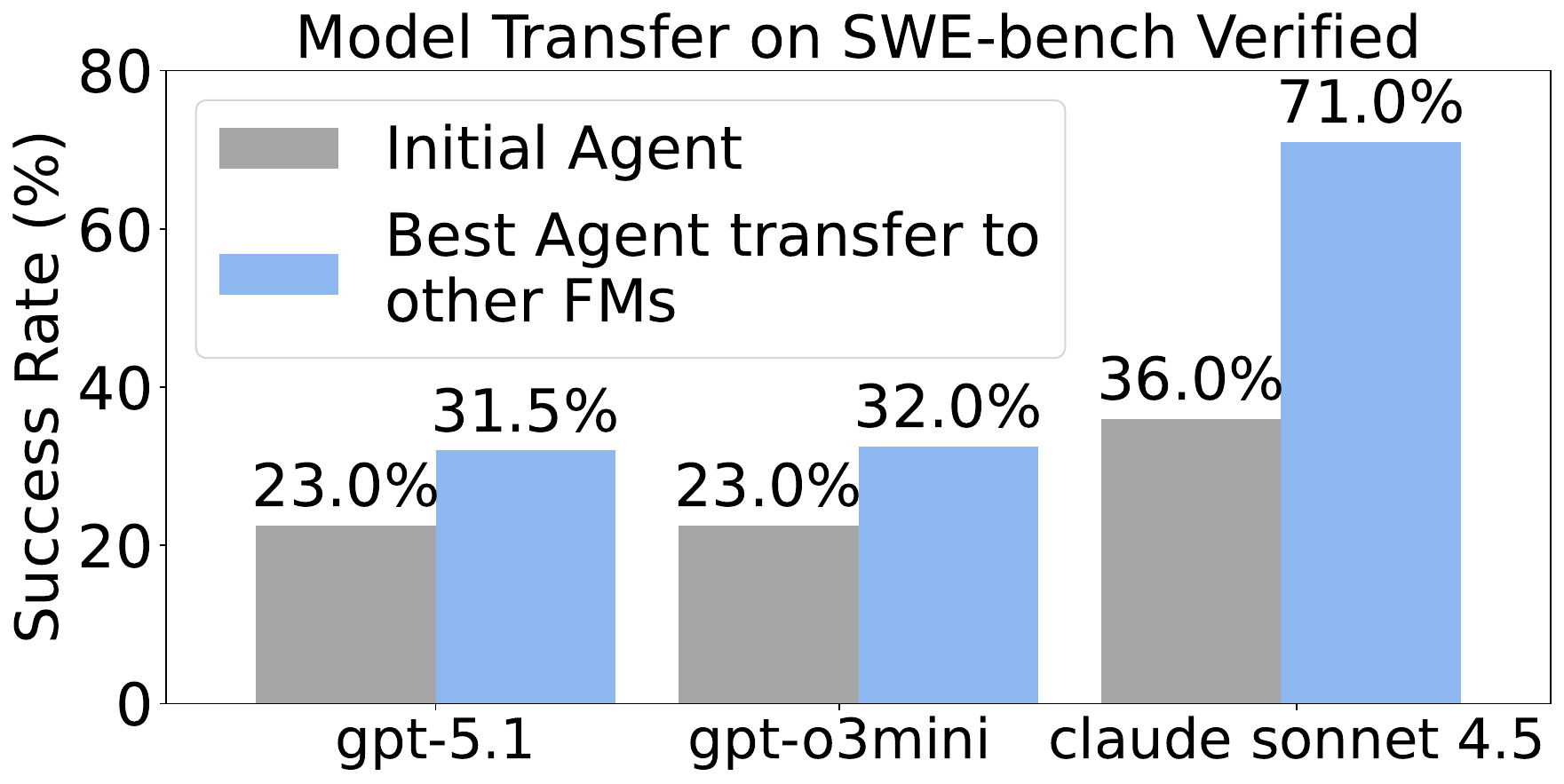}\\[4pt]
  \includegraphics[width=\linewidth]{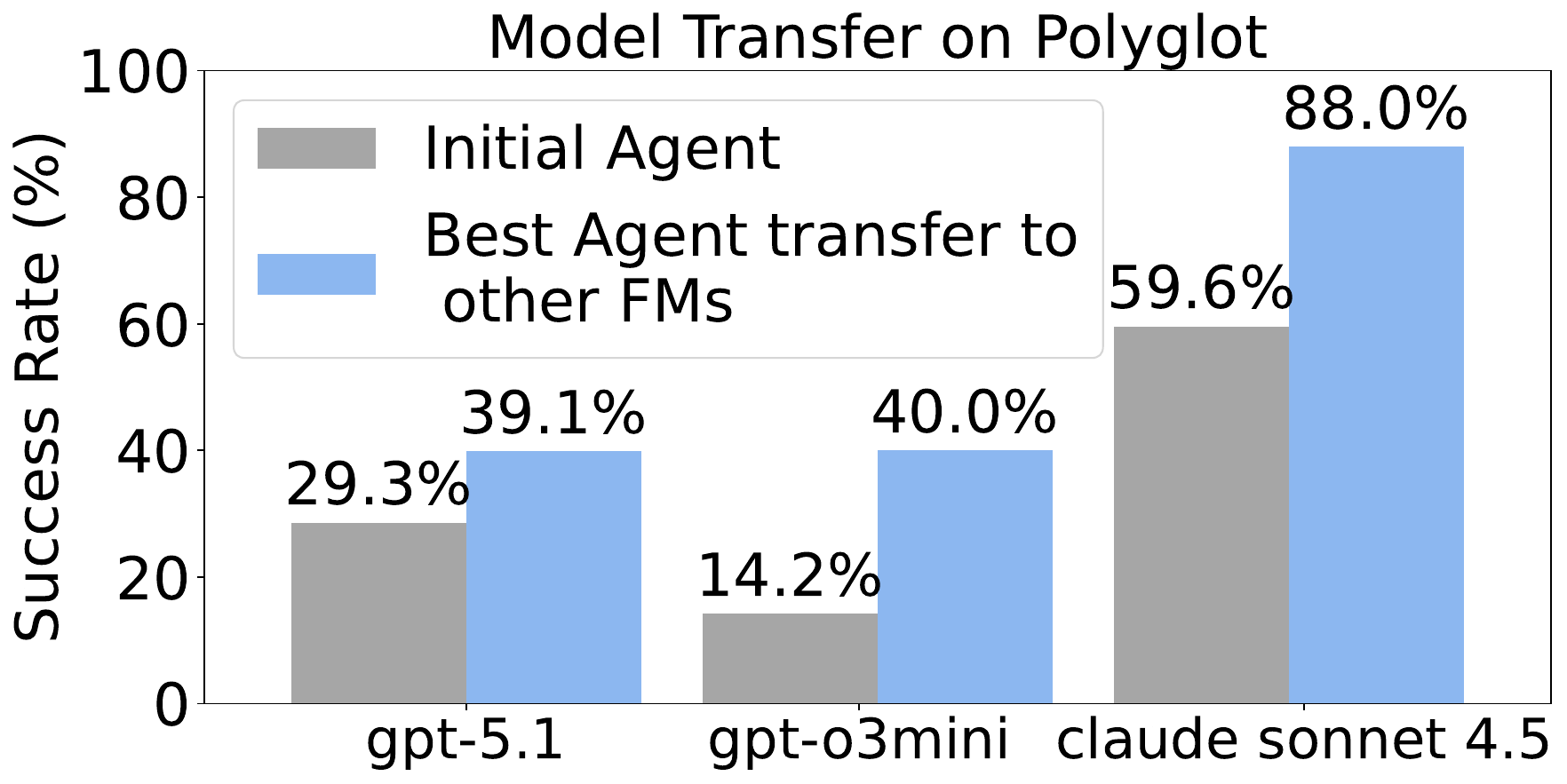}
  \vspace{-6pt}
  \caption{\textbf{Model transfer results on both benchmarks.}
  Across all coding models, the GEA best agent consistently outperforms the corresponding initial (iteration--0) agent,
  demonstrating that the improvements induced by group-level evolution generalize across different underlying model backbones.}
  \label{fig:model-transfer}
  \vspace{-10pt}
\end{wrapfigure}

As illustrated in Figures~\ref{fig:model-transfer}, GEA's best-discovered agent driven by GPT-series and Claude-series models consistently achieve higher performance than their corresponding initial agents on both SWE-bench and Polyglot benchmarks. This indicates that the improvements induced by group-evolving persist across different backbone models.

Further analysis reveals that all performance-improving patches discovered during GEA evolution, including those from the best agent and the top-3 performing agents, primarily target the agent's workflow and tool usage rather than model-specific prompting, details can be found in Table~\ref{tab:top3_agents} in Appendix.

These findings together with Figures~\ref{fig:model-transfer} demonstrate that although GEA leverages a specific backbone model to drive evolution, it discovers agent-level improvements that are largely model-agnostic and the evolved agents could generalize across different coding models.
\begin{table}[t]
    \centering
    \small
    % 增加行高，使表格更透气
    \renewcommand{\arraystretch}{1.2}
    % 调整列间距 (可选)
    \setlength{\tabcolsep}{8pt}
    
    \begin{tabular}{lcccccc}
        \toprule
        \textbf{Method} & E1 & E2 & E3 & E4 & E5 & \textbf{Avg.} $\downarrow$ \\
        \midrule
        DGM & 5  & 4  & 5  & 6  & 5  & 5.0 \\
        GEA (Ours) & 1  & 1  & 2  & 1  & 2  & \textbf{1.4} \\
        \bottomrule
    \end{tabular}
    \vspace{4pt}
    \caption{
    \textbf{Robustness to framework-level bugs.}
    We report the number of evolution iterations required to repair injected bugs across five independent trials (E1--E5). 
    Each entry denotes the number of iterations needed for the agent to successfully repair the injected framework-level bug in a given trial.  \textbf{Lower means} that
    GEA repairs bugs significantly \textbf{faster} than DGM.
    }
    \label{tab:robustness-results}
\end{table}

\subsection{Robustness} To evaluate the robustness of GEA, we introduce framework-level bugs by manually injecting errors into agent implementations. Specifically, we randomly select an agent from the GEA archive and manually inject framework-level bugs into its implementation. We then form a group consisting of this faulty agent and another bug-free agent from its original parent group, and perform group evolution to assess whether GEA can leverage experience from the better-performing agent (i.e., the one without framework-level bugs) to repair the faulty one. For comparison, under the self-evolution setting, the bugged agent evolves independently without access to external experience sharing. In both settings, we measure the number of iterations required to successfully repair the bug.

As shown in Table~\ref{tab:robustness-results}, across five independent trials, GEA requires only 1.4 iterations on average to repair the injected bugs, whereas the self-evolving baseline (DGM) requires 5. This substantial gap demonstrates that group-evolving agents benefit from intra-group experience sharing, enabling successful framework-level experiences from better-performing agents to guide the repair of faulty ones, confirming the robustness of the group-evolving paradigm.
\section{Conclusion}

We introduce \textbf{G}roup-\textbf{E}volving \textbf{A}gents (GEA), a new paradigm for open-ended self-improvement that treats a \emph{group of agents}, rather than an individual agent, as the fundamental unit of evolution.
By enabling explicit experience sharing and reuse within the group, agents can learn from each other's evolutionary experiences and adaptively integrate complementary improvements throughout evolution.

Compared to individual-centric self-evolving approaches, GEA more effectively consolidates valuable exploratory outcomes from early stages into the best-performing agents, efficiently transforming transient diversity into long-term useful experience. 
As a result, group-level evolution achieves substantially stronger performance given the same number of evolved agents.

Further analysis shows that GEA's improvements primarily stem from enhancements to agent workflows and tool usage, rather than overfitting to a specific coding model. Therefore, its gains transfer consistently across different models, including both GPT-series and Claude-series.

In addition, GEA exhibits stronger robustness than individual-centric self-evolving approaches: through group-level experience reuse, better-performing agents can guide the repair of faulty ones, enabling GEA to recover from framework-level bugs with fewer evolution iterations.
\section*{Impact Statement}
GEA demonstrates the potential and viability of group-evolving open-ended systems to autonomously modify their own implementation for continuous improvement. While this potential aligns with the goal of building AI that benefits humanity, open-ended exploration also carries inherent considerations worth noting. For instance, the evolutionary process may inadvertently introduce directions misaligned with human intent while consuming substantial computational resources, or produce patches that lack structural clarity, leading to increasingly complex systems that are difficult to fully understand. Therefore, it is essential to establish appropriate boundaries and guide the system to preserve exploratory diversity while ensuring alignment with human intent. Following~\citet{zhang2025darwin}, all experiments in this work are conducted in isolated sandbox environments, thereby limiting potential impacts on host systems.

On the other hand, although we focus on evolving agents' coding capabilities in this work, this paradigm has broader potential applications, for example, enabling systems to mitigate biases through self-improvement, thereby becoming more trustworthy and beneficial for social good.

\bibliographystyle{unsrtnat}  
\bibliography{ref}

\appendix
\newpage

\section{Appendix}

\subsection{Cost Estimate}
The primary cost for both GEA and DGM arises from benchmark evaluation. Since we generate the same number of agents for both methods, their overall costs are very similar. Following the settings described in Sections~\ref{sec:experimental_settings} and~\ref{sec:baselines_settings}, the estimated cost of completing a full run is approximately USD 13,000 per method on SWE-bench and USD 1,500 on Polyglot. A more detailed estimated cost breakdown is provided below:

\begin{table}[h]
\centering
\begin{tabular}{cccc}
\toprule
\textbf{Coding Model} & \textbf{Benchmark} & \textbf{Number of Tasks} & \textbf{Cost Estimate (USD)} \\
\midrule
Claude Sonnet 4.5 & SWE-bench & 60 & \$370 \\
Claude Sonnet 4.5& Polyglot & 60 & \$60 \\
Claude Haiku 4.5& SWE-bench & 60 & \$120 \\
Claude Haiku 4.5& Polyglot & 60 & \$20 \\
\bottomrule
\end{tabular}
\label{tab:cost}
\end{table}

\subsection{Case Study}
% In preamble (once):
% \usepackage{booktabs}
% \usepackage{tabularx}
% \usepackage{array}

\begin{table}[h]
\centering
\begin{tabular}{@{}llc@{}}
\toprule
\textbf{Agent} & \textbf{Patch Description} & \textbf{$\Delta$ Score} \\
\midrule
\multirow{7}{*}{Top-1} 
& docutils\_debug & +0.10 \\
& Modified core agent logic & +0.10 \\
& Updated tests logic & +0.22 \\
& Add snippet\_extract tool & +0.10 \\
& Updating edit tools (v1) & +0.01 \\
& Updating edit tools (v2) & +0.05 \\
& Updating edit tools (v3) & +0.02 \\
\midrule
\multirow{6}{*}{Top-2} 
& docutils\_debug & +0.10 \\
& Modified core agent logic & +0.10 \\
& Updated tests logic & +0.22 \\
& Add snippet\_extract tool & +0.10 \\
& Add bash tools & +0.01 \\
& Updating edit tools & +0.07 \\
\midrule
\multirow{5}{*}{Top-3} 
& docutils\_debug & +0.10 \\
& Modified core agent logic & +0.10 \\
& Updated tests logic & +0.22 \\
& Add multi-file tools (v1) & +0.01 \\
& Add multi-file tools (v2) & +0.07 \\
\bottomrule
\end{tabular}
\caption{\textbf{Evolutionary trajectories of the top-3 performing agents discovered by GEA on SWE-bench Verified.}
Across all three agents, the performance-improving patches primarily focus on enhancing the agents’ workflows and tool usage, rather than relying on model-specific prompting strategies.}
\label{tab:top3_agents}
\end{table}

\end{document}

\maketitle

\begin{abstract}
This supplementary document provides additional details, analyses, and experimental evidence supporting the proposed DMLR framework. Section~\ref{a} presents the full evaluation protocol, including dataset descriptions, unified prompting configurations, baseline implementations, and the procedures used for visual-dependency, confidence, faithfulness, and hallucination studies. Section~\ref{b} reports a comprehensive set of hyperparameter choices and ablation settings used to assess the robustness of latent reasoning and dynamic visual injection. Section~\ref{c} includes extended quantitative results such as Pass@k evaluations and comparisons against training-based reasoning methods. Section~\ref{d} provides theoretical analyses, including proofs of confidence–quality gradient alignment, boundedness of the positive-correlation region, and the formal justification for why dynamic visual injection improves confidence. Section~\ref{e} further elaborates on the design choices, mechanisms, and stability analyses of the Dynamic Visual Injection module. Section~\ref{f} presents qualitative case studies demonstrating improvements in perception quality, reasoning reliability, and reasoning conciseness under DMLR. Section~\ref{g} concludes with the complete set of prompts used across all experiments, ensuring transparency and reproducibility. We will make all code publicly available in future releases.
\end{abstract}

\appendix

\section*{Appendix Table of Contents}
\startcontents[sections]
\printcontents[sections]{l}{1}{\setcounter{tocdepth}{2}}

\clearpage
\renewcommand{\thesubsection}{\thesection\arabic{subsection}}

\section{More Detailed about Evaluation}
\label{a}

\subsection{Datasets}
\begin{itemize}[leftmargin=*]
\item \textbf{MathVista$_{\text{mini}}$} is a benchmark for mathematical reasoning in visual contexts, aggregating diverse multimodal math tasks that require fine-grained visual understanding and compositional numerical reasoning.

\item \textbf{MathVision$_{\text{mini}}$} is a curated benchmark of competition-level visual math problems spanning multiple disciplines and difficulty levels to assess multimodal models’ mathematical reasoning under challenging and diverse settings.

\item \textbf{MM Math} is a benchmark of open-ended math problems with visual contexts that supports both outcome and process evaluation, enabling detailed analysis of multimodal reasoning behaviors and typical error patterns.

\item \textbf{HallusionBench} is a benchmark for image-context reasoning that uses carefully structured question pairs to diagnose hallucination, visual illusion, and logical inconsistency in large vision-language models.

\item \textbf{MMVP} is a benchmark built from multimodal visual patterns designed to expose “CLIP-blind’’ image–text pairs, revealing systematic visual perception failures and hallucinated explanations in multimodal LLMs. 

\item \textbf{MMStar} is a vision-indispensable multimodal benchmark composed of carefully human-filtered samples that ensure true visual dependency while evaluating core multimodal capabilities along multiple fine-grained axes.

\item \textbf{ScienceQA} is a multimodal multiple-choice science benchmark with rich textual and visual contexts, lectures, and explanations that spans diverse subjects and skills, supporting evaluation of both answer accuracy and explanation quality.
\end{itemize}

For all datasets, we limit the maximum sample size to 1000 instances.

\subsection{Evaluation Setting} 
We adopt a unified prompting setup for all models. Unless otherwise stated, we use greedy decoding (\texttt{do\_sample=False}) for all generation tasks.

\paragraph{System Prompt.}
\begin{center}
\fbox{\parbox{0.96\linewidth}{
\small
A conversation between User and Assistant. The user asks a question, and the Assistant solves it. The assistant first thinks about the reasoning process in the mind and then provides the user with the answer. The reasoning process and answer are enclosed within \texttt{<think>} \texttt{</think>} and \texttt{<answer>} \texttt{</answer>} tags, respectively, i.e., \texttt{<think>} reasoning process here \texttt{</think>} \texttt{<answer>} answer here \texttt{</answer>}.
}}
\end{center}

\paragraph{Task Prompt.}
\begin{center}
\fbox{\parbox{0.96\linewidth}{
\small
Please analyze the image carefully and solve this problem step by step. Show your reasoning process clearly, then put your final answer within \texttt{\textbackslash boxed\{\}}.\\[4pt]
\textbf{Question:} \emph{[Problem Text]}
}}
\end{center}

For all benchmarks considered in our experiments, the ground-truth answers are verifiable; we use regular expressions to extract the content within \texttt{\textbackslash boxed\{\}} from the model outputs and then match it against the correct answers.

\subsection{Baseline Method}

\begin{itemize}[leftmargin=*] \item \textbf{CCoT}~\cite{Mitra_2024_CVPR} (Compositional Chain-of-Thought) is a zero-shot prompting method that utilizes scene graphs to extract compositional knowledge. It first generates a scene graph to capture object attributes and relationships, which is then used to guide the LMM in generating the final response without requiring fine-tuning.

\item \textbf{ICoT}~\cite{gao2025interleavedmodalchainofthought} (Interleaved-modal Chain-of-Thought) generates sequential reasoning steps consisting of paired visual and textual rationales. It employs a plug-and-play Attention-driven Selection (ADS) mechanism to dynamically identify and insert relevant image regions into the reasoning chain based on the model's attention maps.

\item \textbf{Multimodal-CoT}~\cite{zhang2024multimodalchainofthoughtreasoninglanguage} incorporates language and vision modalities into a two-stage framework that separates rationale generation from answer inference. It leverages multimodal information to generate intermediate reasoning chains, which subsequently serve as the basis for inferring the final answer, effectively mitigating hallucination. 

\item\textbf{MCOUT (Training)}~\cite{pham2025multimodalchaincontinuousthought} is a latent-space reasoning framework that replaces traditional text-based CoT with continuous hidden-state “thought vectors,” enabling models to reason directly in a unified multimodal latent space.

\item \textbf{IVT-LR (Training) }~\cite{chen2025reasoningdarkinterleavedvisiontext} is a new multimodal reasoning method that performs all intermediate thinking directly in the latent space by combining hidden textual states with dynamically selected visual embeddings.
\end{itemize}

\subsection{Additional explanations for the validation experiments in Section 3.}

\textbf{Visual Dependency Analysis.} \textit{\textbf{(1) Robustness Across Multiple Perturbation Types.}} To ensure that the visual-dependency metric is not influenced by specific perturbation noise, we evaluate four representative reasoning models, R1-OneVision, VLAA-Thinking, Ocean-R1, and ThinkLite-VL, across two reasoning datasets (MathVision and MathVista) and two perception datasets (MMVP and HallucinationBench). Instead of relying on a single distortion strategy, we apply four distinct perturbation types including block occlusion, color jitter, random region masking, and Gaussian blur. Although these perturbations substantially differ in the way they alter local textures or global color statistics, the resulting visual-dependency distributions remain highly consistent across all models and datasets, as shown in Table~\ref{tab:perturb-consistency}. This consistency confirms that the sparsity pattern observed in the main paper is not tied to any specific perturbation method but reflects a stable and intrinsic property of multimodal reasoning models.

\begin{table}[h]
\centering
\caption{Consistency of visual-dependency distributions across models, datasets, and perturbation types. 
Values represent the average Pearson correlation between dependency curves under different perturbations. Higher is better.}
\vspace{0.15cm}
\resizebox{\linewidth}{!}{
\begin{tabular}{lcccc}
\toprule
\textbf{Model} & \textbf{MathVision} & \textbf{MathVista} & \textbf{MMVP} & \textbf{HallucinationBench} \\
\midrule
R1-OneVision & 0.92 & 0.89 & 0.94 & 0.91 \\
VLAA-Thinking & 0.90 & 0.88 & 0.93 & 0.90 \\
Ocean-R1 & 0.87 & 0.85 & 0.91 & 0.88 \\
ThinkLite-VL & 0.89 & 0.86 & 0.92 & 0.89 \\
\midrule
\textbf{Average} & \textbf{0.90} & \textbf{0.87} & \textbf{0.93} & \textbf{0.90} \\
\bottomrule
\end{tabular}}
\label{tab:perturb-consistency}
\end{table}

\noindent \textit{\textbf{(2) Noise Reduction and Token-Level Aggregation.}} To obtain a stable estimation of token-level visual dependency, each dependency value is averaged across five independently perturbed versions of the same image, effectively reducing stochastic fluctuations associated with any single perturbation. Since reasoning chains generated by different models vary in length, direct position-wise averaging would introduce structural bias. To avoid this, we adopt a relative-position alignment scheme that normalizes each reasoning chain to a comparable relative index space. Additionally, the dependency value of each chain is summarized using the median rather than the mean, preventing rare but extreme perturbation-induced peaks from affecting the aggregated distribution. These methodological choices ensure that the resulting dependency curves robustly reflect the inherent behavior of the models rather than noise introduced during perturbation.

\noindent \textit{(3) \textbf{Semantic Interpretation of High-Dependency Regions.} }Beyond statistical validation, we analyze the semantic roles of tokens exhibiting high visual dependency. Across all datasets and models, we observe that such tokens consistently align with reasoning stages in which visual grounding is intrinsically required (as shown in Table~\ref{tab:semantic-high-dependency}), such as the initial image description, spatial or numerical verification, and operations involving visual alignment (e.g., comparing, locating, or counting). These consistent semantic correspondences demonstrate that the detected peaks in visual dependency are not random artifacts but meaningful signals of when the model must rely on perceptual evidence to proceed. This structured, task-aligned sparsity further validates that multimodal LLMs rely on vision only at key points during reasoning, rather than uniformly throughout the reasoning chain.

\begin{table}[h]
\centering
\caption{
Distribution of high visual-dependency tokens across semantic reasoning stages.
Values indicate the percentage of top-10\% dependency tokens falling into each stage.
}
\vspace{0.15cm}
\resizebox{\linewidth}{!}{
\begin{tabular}{lcc}
\toprule
\textbf{Semantic Stage} & \textbf{MathVision} & \textbf{MMVP} \\
\midrule
Initial image description & 38.2\% & 36.5\% \\
Spatial or numerical verification & 27.4\% & 30.1\% \\
Visual alignment (compare / locate / count) & 24.7\% & 26.9\% \\
Non-visual reasoning steps & 9.7\% & 6.5\% \\
\bottomrule
\end{tabular}}
\label{tab:semantic-high-dependency}
\end{table}

\noindent \textbf{Confidence Analysis.} \textit{(1) \textbf{Experimental Setting for Correct and Incorrect Reasoning Chains.}} To reliably distinguish correct from incorrect reasoning chains, we adopt a controlled evaluation protocol that ensures accuracy labels are consistent and free from formatting artifacts. For each example, the model generates multiple reasoning chains along with a final predicted answer, and the correctness of each chain is determined by comparing the model’s final answer to the ground-truth label after normalizing answer formats to eliminate superficial mismatches. To prevent cases in which a chain produces the correct answer but exhibits logically incoherent reasoning from being misclassified, we further employ GPT-4o as an external judge. The judge is provided with the question, the full chain, and the final answer, and is instructed to verify whether the reasoning is logically consistent and legitimately leads to the given answer. All judgments are performed with zero temperature to maintain high determinism. Correct and incorrect chains are then sampled and balanced across datasets, and spot-checked manually to ensure labeling quality. This procedure provides a robust foundation for analyzing confidence patterns between truly correct and genuinely incorrect reasoning chains.

\noindent \textit{(2) \textbf{Experimental Setting for Faithful and Unfaithful Reasoning Chains.}} To assess the faithfulness of model-generated reasoning, we evaluate whether each chain genuinely relies on the evidence provided in the question and image, rather than resorting to template-based explanations, unsupported assumptions, or coincidental correctness. Each chain is reviewed by GPT-4o, which receives the question, the full reasoning steps, and either the original image or a standardized image description. The model judge is explicitly instructed to determine whether the reasoning aligns with the available evidence and whether each step contributes meaningfully to the final answer. Because reasoning faithfulness is often more subtle than answer correctness, we introduce a dual-evaluation protocol in which each chain is assessed twice independently, and majority voting is used to resolve any discrepancies. Chains that arrive at the correct answer through unsupported or spurious reasoning are deliberately marked as unfaithful to avoid conflating guessing with faithful inference. Samples are then balanced across datasets to ensure that faithfulness-based confidence analyses reflect structural differences in reasoning quality rather than dataset biases.

\noindent \textit{(3) \textbf{Experimental Setting for Hallucination and Non-hallucination Reasoning}} To study the relationship between confidence dynamics and visual grounding, we focus on the subset of reasoning steps that explicitly reference visual content. From each chain, we automatically extract statements mentioning visual entities, attributes, or spatial relationships. These statements, together with the corresponding image or image description, are presented to GPT-4o, which evaluates whether each statement is factually supported by the visual evidence or constitutes a hallucination. The judge is instructed to assess visual consistency rather than linguistic fluency, and to treat any fabricated or contradictory visual claims as hallucinations. To improve reliability, each statement is evaluated twice under zero temperature, and samples with inconsistent labels undergo manual verification. After labeling, each hallucinated or grounded statement is aligned with the model’s token-level confidence at the corresponding reasoning step. By comparing the confidence distributions of hallucinated versus non-hallucinated statements, we reveal systematic patterns showing that hallucinated visual steps tend to exhibit lower confidence and higher uncertainty. This experimental design thus provides strong empirical support for the link between confidence signals and visual grounding quality.

\noindent All prompts used in these experimental settings are provided in the Appendix~\ref{prompt} for completeness and reproducibility.

\section{More Detailed about Parameter Analyses}
\label{b}
\subsection{Parameters Setup}

\begin{itemize}[leftmargin=*]

\item \textbf{Latent Think Tokens $\mathcal{T}$:} We set the number of latent think tokens to 4. During generation, after each latent token the model dynamically injects a visual patch into the latent stream to refresh its internal perception state.

\item \textbf{Image Patches $m$:} We dynamically insert visual patches into the latent stream. At initialization, we inject 2 patches; at each subsequent iteration, we select $m = 2$ patches with the highest attention scores and append them after each latent think token, with at most 16 patches inserted per iteration. Additionally, we set the image processor's max pixel size to 256 for all inputs.

\item \textbf{Optimization Parameters:} We perform 15 latent optimization steps with a learning rate of $1\times 10^{-3}$. To ensure stable exploration in the latent space, the perturbation magnitude is set to $\sigma = 0.1$ (10\%) with a decay factor of 0.95, and we fix the random seed to 42 for all experiments.

\item \textbf{Environment Setup:} All experiments are conducted on NVIDIA H100 GPUs. For a fair comparison, we run all baselines and our method in \texttt{float32} precision and use the \texttt{eager} attention backend for inference.
\end{itemize}

\subsection{Ablation Experimental Setup}

For the ablation studies presented in Sec.~5.3, we maintain the same parameter settings as described above. We employ VLAA-Thinker-Qwen2.5VL-7B as the reasoning model and Qwen2.5-VL-7B-Instruct as the non-reasoning model. To ensure efficiency, all ablation experiments are conducted on a subset of 300 samples randomly selected from each dataset.

\section{More Results}
\label{c}
\textbf{Pass@\texorpdfstring{$k$}{k} Evaluation.} We employ the Pass@$k$ metric to evaluate the accuracy of the model's generated answers. Following standard practice, we calculate the unbiased estimator using the formula:
\begin{equation}
    \text{Pass}@k = 1 - \frac{\binom{n - c}{k}}{\binom{n}{k}},
\label{eq1}
\end{equation}
where $n$ is the total number of sampled solutions per problem, $c$ is the number of correct samples, and $k$ is the budget of samples selected. Note that when $k=1$, this simplifies to the empirical accuracy $\text{Pass}@1 = \frac{c}{n}$. 

Using the unbiased estimator of Pass@\emph{k} defined in Equation~\ref{eq1}, we evaluate
all methods under a sampling budget of $k=8$. This metric measures the
probability that at least one of the sampled solutions is correct, offering a
more reliable estimate of reasoning robustness compared to single-sample
accuracy. As shown in Table~\ref{pass}, integrating DMLR consistently
improves Pass@8 across mathematics reasoning, visual reasoning, and multimodal
composition tasks. For the stronger Qwen2.5-VL-7B backbone, DMLR achieves gains ranging from
+2.4\% to +4.8\%, with the largest improvements observed on MM-Math,
HallusionBench, and MMVP benchmarks that demand multi-step reasoning or precise
visual grounding. Even with the smaller 3B backbone, where reasoning capacity is
more limited, DMLR maintains steady improvements of +1.0\% to +4.8\%,
indicating that the proposed dynamic latent reasoning process effectively
enhances solution diversity and increases the likelihood of producing at least
one correct prediction within the budgeted $k$ samples. Overall, the consistent uplift in Pass@8 demonstrates that DMLR not only refines
the latent reasoning trajectory but also mitigates confidently incorrect
generations, leading to stronger performance across all evaluated benchmarks.

\noindent \textbf{Comparison with Training-Based Methods.} Table~\ref{additional} provides a comprehensive comparison between
DMLR and two additional training-based reasoning enhancement methods, namely
MCOUT and IVT-LR. These approaches enforce implicit reasoning structures through
explicit training objectives: MCOUT encourages the model to internalize stable
multi-step reasoning patterns via supervised signals, whereas IVT-LR injects
learned latent guidance vectors to impose structured constraints on the
model’s internal reasoning trajectory. Both methods require additional data,
parameter updates, and dedicated training phases designed to encode specific
implicit patterns into the model’s latent space.

Across all backbone models and task categories, DMLR consistently achieves
substantial performance gains, frequently surpassing these training-based
approaches despite being entirely training-free. On Qwen2.5-VL-7B, DMLR
outperforms IVT-LR on several challenging benchmarks—including MathVision,
MM-Math, HallusionBench, and MMStar—demonstrating that dynamic latent reasoning
refinement can effectively enhance model performance without modifying model
parameters. Similar improvements are observed on Qwen2.5-VL-3B, LLaVA-1.6-7B,
and Qwen3-VL variants, with DMLR exceeding the strongest training-based
baselines by 1.2\%-5.0\%. Notably, in lighter-weight models, where training-based methods are constrained by limited model capacity and potential
overfitting, DMLR exhibits superior generalization due to its adaptive inference
mechanism.

These findings indicate that DMLR offers advantages beyond the realm of
training-free inference. Its plug-and-play nature is complemented by a dynamic
latent optimization process that adjusts the internal reasoning trajectory in
real time, enabling task-adaptive refinement that static training-based patterns
cannot provide. In contrast, the implicit structures learned through training
are fixed after optimization and thus may generalize less effectively across
tasks or distributional shifts.

It is important to note that we do not include “think-with-image” models such as
DeepEye~\cite{zheng2025deepeyesincentivizingthinkingimages} or GRIT~\cite{fan2025gritteachingmllmsthink} in this comparison. These approaches rely on extensive
multimodal pretraining with large-scale curated or synthetic datasets to acquire
explicit or implicit image-conditioned reasoning pathways. Their performance
benefits primarily arise from substantial additional data and intensive training
procedures rather than inference-time optimization. As this work focuses on
evaluating latent-state refinement \emph{without any additional training}, such
methods fall outside the scope of this comparison.

\begin{table*}[h]
\centering
\caption{Comparison of different reasoning methods and DMLR across various benchmarks. All metrics are reported in Accuracy (\%).}
\vspace{-0.2cm}
\large
\renewcommand{\arraystretch}{1.05}
\resizebox{0.9\linewidth}{!}{%
\setlength{\tabcolsep}{3pt}
\begin{tabular}{ll|ccc|cc|cc}
\toprule[1.5pt]
\multicolumn{2}{c}{} &
\multicolumn{3}{c}{\textbf{Mathematics Reasoning\uparrow} 
} &
\multicolumn{2}{c}{\textbf{Visual Reasoning\uparrow}} &
\multicolumn{2}{c}{\textbf{Multimodal Composition\uparrow}} \\
\cmidrule(lr){3-5} \cmidrule(lr){6-7} \cmidrule(lr){8-9}
\textbf{Method} & \textbf{Model} &
MathVista$_{\text{mini}}$ &
MathVision$_{\text{mini}}$ &
MM-Math &
HallusionBench &
MMVP & MMStar &
ScienceQA \\
\midrule

Vanilla &  & 58.7 & 21.6 & 37.5 & 65.4 & 68.7 & 59.3 & 49.7 \\
Multimodal COT &  & 56.4 & 21.8 & 35.6 & 63.6 & 68.1 & 57.9 & 49.5  \\
CCOT &  & 57.8 & 22.5 & 36.3 & 64.9 & 69.0 & 58.7 & 50.2 \\
ICoT &  & 58.9 & 23.3 & 37.0 & 65.5 & 69.3 & 60.4 & 50.4\\
MCOUT  &  & 58.0  & 22.7 & 37.5 & 65.0 & 69.4 & 59.2 & 50.9 \\
IVT-LR  &  &  58.5 & 23.5  & 38.0  & 65.2  & 69.8  & 58.4  & 49.8  \\

\rowcolor{lightblue}
\textbf{\textit{+DMLR (Ours)}} & 
\multirow[c]{-5}{*}[1pt]{\rotatebox{90}{\parbox{2.5cm}{\centering \fontsize{11pt}{13pt}\selectfont Qwen2.5 \\VL 7B}}}
 & \textbf{59.1} \up{0.40\%} & \textbf{24.4} \up{2.8\%} & 38.8 \up{1.3\%} & 65.8 \up{0.4\%} & \textbf{70.1} \up{1.4\%} & \textbf{60.1} \up{0.8\%} & \textbf{51.3} \up{1.6\%} \\
\midrule\midrule

Vanilla &  & 48.2 & 15.7 & 29.0 & 64.2 & 55.6 & 50.2 & 44.1 \\
Multimodal COT &  & 47.3 & 14.3 & 28.5 & 63.8 & 54.4 & 48.5 & 42.9 \\
CCOT & & 48.0 & 15.6 & 30.2 & 64.0 & 55.5 & 49.3 & 44.5 \\
ICoT &  & 49.8 & 16.0 & 30.6 & 64.7 & 55.9 & 49.6 & 45.0 \\
MCOUT   &  & 49.4 & 15.8 & 32.3  & 63.0 & 55.6  & 50.2 & 45.5 \\
IVT-LR  &   & 50.2  & 16.4 & 32.0  & 63.4  & 55.9 & 50.3 & 46.0  \\
\rowcolor{lightblue}
\textbf{\textit{+DMLR (Ours)}} &
\multirow[c]{-5}{*}[1pt]{\rotatebox{90}{\parbox{2.5cm}{\centering \fontsize{11pt}{13pt}\selectfont Qwen2.5 \\VL 3B}}}
 & \textbf{51.0} \up{2.80\%} & \textbf{17.7} \up{2.74\%} & \textbf{33.3} \up{4.3\%} & \textbf{64.7} \up{0.5\%} & \textbf{56.8} \up{1.26\%} & \textbf{51.2} \up{1.00\%} & \textbf{46.9} \up{2.8\%}\\
\midrule\midrule

Vanilla &  & 61.1 & 23.5 & 41.5 & 62.0 & 68.3 & 58.9 & 50.6 \\\
Multimodal COT &  & 59.6 & 23.1 & 40.6 & 62.8  & 67.2 & 57.1 & 48.2  \\
CCOT &  & 60.5  & 24.8  & 41.8  & 64.6 & 68.0 & 59.0 & 49.4\\
ICoT &  & 61.4 & 25.0 & 42.3 & 65.9 & 68.3 & 58.2 & 50.6 \\
MCOUT   &  & 60.5  & 26.3  & 41.9 & 66.0 & 68.5 & 58.5  & 49.9  \\
IVT-LR  &   & 61.4   & 26.6 &  42.5  & 66.7  & 68.9 & 59.0 & 50.3 \\
\rowcolor{lightblue}
\textbf{\textit{+DMLR (Ours)}} &
\multirow[c]{-4}{*}[1pt]{\rotatebox{90}{\parbox{1.0cm}{\centering \fontsize{11pt}{13pt}\selectfont VLAA \\ Thinking 7B}}}
 & \textbf{62.9} \up{1.80\%} & \textbf{27.6} \up{4.10\%} & \textbf{43.9} \up{2.41\%} & \textbf{67.9} \up{5.94\%} & \textbf{69.4} \up{1.1\%} & \textbf{59.2} \up{0.3\%} & \textbf{51.98} \up{1.38\%} \\
\midrule\midrule

Vanilla &  & 51.2 & 18.7 & 40.7 & 62.1 & 67.0 & 52.1 & 50.9 \\
Multimodal COT &  & 52.5 & 18.9 & 39.6 & 62.5 & 68.0 &  51.6 & 51.7 \\
CCOT &  & 53.4 & 20.3 & 40.8 & 63.0 & 68.9 & 53.5 & 52.8 \\
ICoT &  & 55.6  & 21.5 & 41.7 & 63.8 & 69.6 & 54.0 & 54.4 \\
MCOUT   &  & 56.3  &  21.6  & 40.9  & 63.3   & 69.9 & 55.0 & 53.8 \\
IVT-LR  &  & 57.2  &  22.3 & 41.1 & 64.5 & 70.4  & 55.9 & 54.8 \\
\rowcolor{lightblue}
\textbf{\textit{+DMLR (Ours)}} &
\multirow[c]{-4}{*}[1pt]{\rotatebox{90}{\parbox{1.2cm}{\centering \fontsize{11pt}{11pt}\selectfont \quad R1 \\ OneVision \quad \quad 7B}}}
 & \textbf{58.0} \up{6.81\%} & \textbf{23.3} \up{4.56\%} & \textbf{42.9} \up{2.21\%} & \textbf{64.1} \up{2.09\%} & \textbf{71.9} \up{4.93\%} & \textbf{56.2} \up{4.14\%} & \textbf{55.4} \up{4.52\%}\\
\midrule\midrule

Vanilla &  & 66.0 & 32.9 & 66.2 & 73.2 & 71.9 & 68.1 & 54.1 \\
Multimodal COT &  & 64.8 & 32.8 & 65.1 & 73.0 & 69.6 & 66.9  &  53.2 \\
CCOT &   & 66.5 & 33.3 & 65.5 & 73.5 & 70.3 & 68.8 & 54.4 \\
ICoT &  & 66.2 & 34.9 & 66.8 & 74.5 & 71.8 & 69.3 & 55.8 \\
MCOUT   & & 65.5 &  33.8 & 65.0 & 73.7 & 72.0 & 68.5  & 54.3  \\
IVT-LR  &  & 65.9   & 34.5 & 66.1 & 73.4  & 71.5 & 69.4 & 54.8  \\
\rowcolor{lightblue}
\textbf{\textit{+DMLR (Ours)}} &
\multirow[c]{-5}{*}[1pt]{\rotatebox{90}{\parbox{2.5cm}{\centering \fontsize{11pt}{13pt}\selectfont Qwen3 \\VL 8B}}}
 & \textbf{66.9} \up{0.9\%} & \textbf{36.2} \up{3.34\%}& \textbf{67.7} \up{1.51\%} & \textbf{74.6} \up{1.48\%} & \textbf{72.8} \up{0.97\%}& \textbf{70.0} \up{1.91\%} & \textbf{55.6} \up{1.48\%} \\
\midrule\midrule

Vanilla &  & 64.7 & 24.2 & 65.4 & 71.6 & 71.3 & 57.4 & 52.4 \\
Multimodal COT &  & 62.3 & 24.8 & 63.9 & 70.0 & 69.6 & 57.7 & 53.0  \\
CCOT &  & 64.5 &  26.6 & 64.8 & 71.5 & 71.2 & 58.8 & 52.9 \\
ICoT &  & 64.5 & 27.5 & 65.0 & 72.2 & 72.5 & 59.3 & 53.7 \\
MCOUT   & &  63.6  & 28.0  &  64.5  & 70.1 & 72.0 & 59.6 &  53.9  \\
IVT-LR  &  & 64.0  & 28.4  & 64.8  & 71.3 & 71.9 & 58.4  & 53.6 \\
\rowcolor{lightblue}
\textbf{\textit{+DMLR (Ours)}} &
\multirow[c]{-5}{*}[1pt]{\rotatebox{90}{\parbox{2.5cm}{\centering \fontsize{11pt}{13pt}\selectfont Qwen3 \\VL 4B}}}
 & \textbf{65.6} \up{0.93\%} & \textbf{29.4} \up{5.20\%}  & \textbf{65.9} \up{0.5\%} & \textbf{72.7} \up{1.12\%} & \textbf{72.3} \up{0.98\%}& \textbf{60.3} \up{2.88\%} & \textbf{54.9} \up{2.48\%} \\
\bottomrule[1.5pt]
\end{tabular}
}
\vspace{-0.5em}
\label{additional}
\end{table*}

\begin{table*}[t]
\centering
\caption{Comparison of different reasoning methods and DMLR across various benchmarks. 
All metrics are reported as pass@8 (\%). Results are evaluated over a diverse suite of mathematics reasoning, visual reasoning, and multimodal composition tasks under Qwen2.5-VL backbones (7B and 3B).}
\vspace{-0.2cm}
\large
\renewcommand{\arraystretch}{1.05}
\resizebox{0.85\linewidth}{!}{%
\setlength{\tabcolsep}{3pt}
\begin{tabular}{ll|ccc|cc|cc}
\toprule[1.5pt]
\multicolumn{2}{c}{} &
\multicolumn{3}{c}{\textbf{Mathematics Reasoning\uparrow}} &
\multicolumn{2}{c}{\textbf{Visual Reasoning\uparrow}} &
\multicolumn{2}{c}{\textbf{Multimodal Composition\uparrow}} \\
\cmidrule(lr){3-5} \cmidrule(lr){6-7} \cmidrule(lr){8-9}
\textbf{Method} & \textbf{Model} &
MathVista$_{\text{mini}}$ &
MathVision$_{\text{mini}}$ &
MM-Math &
HallusionBench &
MMVP & MMStar &
ScienceQA \\
\midrule

Vanilla &  & 61.2 & 24.3 & 40.1 & 67.9 & 69.1 & 62.4 & 53.3 \\
Multimodal COT &  & 59.4 & 25.8 & 42.2 & 68.3 & 72.5 & 65.2 & 52.9  \\
CCOT &  & 64.4 & 25.1 & 41.7 & 68.5 & 71.8 & 64.7 & 54.1 \\
ICoT &  & 64.1 & 23.9 & 43.6 & 68.9 & 73.3 & 63.7 & 53.6\\
\rowcolor{lightblue}
\textbf{\textit{+DMLR (Ours)}} & 
\multirow[c]{-5}{*}[1pt]{\rotatebox{90}{\parbox{2.5cm}{\centering \fontsize{11pt}{13pt}\selectfont Qwen2.5 \\VL 7B}}}
 & \textbf{65.3} \up{4.1\%} & \textbf{26.7} \up{2.4\%} & \textbf{44.1} \up{4.0\%} & \textbf{70.4} \up{3.5\%} & \textbf{73.9} \up{4.8\%} & \textbf{67.1} \up{4.7\%} & \textbf{58.4} \up{5.1\%} \\
\midrule\midrule

% ---------------- Qwen2.5-VL 3B ----------------
Vanilla &  & 50.5 & 18.2 & 33.4 & 67.1 & 57.9 & 52.7 & 45.9 \\
Multimodal COT &  & 52.2 & 19.4 & 33.0 & 66.9 & 57.5 & 52.9 & 46.3 \\
CCOT & & 54.3 & 19.7 & 35.1 & 66.5 & 58.6 & 53.3 & 46.1 \\
ICoT &  & 54.0 & 19.3 & 34.6 & 67.3 & 58.4 & 53.0 & 46.7 \\
\rowcolor{lightblue}
\textbf{\textit{+DMLR (Ours)}} &
\multirow[c]{-5}{*}[1pt]{\rotatebox{90}{\parbox{2.5cm}{\centering \fontsize{11pt}{13pt}\selectfont Qwen2.5 \\VL 3B}}}
 & \textbf{55.3} \up{4.8\%} & \textbf{20.9} \up{2.7\%} & \textbf{36.3} \up{2.9\%} & \textbf{69.5} \up{2.4\%} & \textbf{59.2} \up{1.3\%} & \textbf{53.7} \up{1.00\%} & \textbf{48.1} \up{2.2\%}\\
\bottomrule[1.5pt]
\end{tabular}
}
\vspace{-0.5em}
\label{pass}
\end{table*}

\section{Theoretical Analyses}
\label{d}
\subsection{Confidence and Quality Gradient Alignment.}

\subsubsection{Problem Setup and Assumptions}
\noindent We consider the latent reasoning state $h \in \mathbb{R}^d$ and two scalar functions:
a confidence function $C : \mathbb{R}^d \to \mathbb{R}$ and a reasoning-quality function
$Q : \mathbb{R}^d \to \mathbb{R}$. During test-time optimization, DMLR updates the latent
state by ascending the confidence landscape:
\begin{equation}
    h^{+} = h + \eta \nabla C(h),
    \quad \eta > 0.
    \label{eq:dmlr_update_conf}
\end{equation}

\paragraph{Assumption A.1 (Differentiability).}
The functions $C$ and $Q$ are twice continuously differentiable on an open set 
$\mathcal{D} \subset \mathbb{R}^d$.
\vspace{-0.2cm}
\paragraph{Assumption A.2 (Smoothness of $Q$).}
There exists a constant $L_Q > 0$ such that
\begin{equation}
    \|\nabla Q(h_1) - \nabla Q(h_2)\|
    \;\le\;
    L_Q \,\|h_1 - h_2\|,
    \quad \forall\, h_1, h_2 \in \mathcal{D}.
    \label{eq:L_smooth_Q}
\end{equation}
\vspace{-0.8cm}
\paragraph{Assumption A.3 (Local Gradient Alignment).}
At a given point $h \in \mathcal{D}$, we have
\begin{equation}
    \nabla C(h)\cdot\nabla Q(h) > 0,
    \qquad
    \nabla C(h) \neq 0.
    \label{eq:local_alignment}
\end{equation}

\subsubsection{Auxiliary Inequality for $L_Q$--Smooth Functions}
\begin{lemma}[Descent lemma form]
\label{lem:descent_lemma}
Under Assumptions~A.1 and~A.2, for any $h \in \mathcal{D}$ and any vector 
$\Delta h$ such that $h + \Delta h \in \mathcal{D}$, we have
\begin{equation}
    Q(h + \Delta h)
    \;\ge\;
    Q(h) + \nabla Q(h)\cdot \Delta h
    - \frac{L_Q}{2}\,\|\Delta h\|^2.
    \label{eq:descent_lemma}
\end{equation}
\end{lemma}

\begin{proof}
Define a scalar function $g : [0,1] \to \mathbb{R}$ by
\begin{equation}
    g(t) = Q(h + t \Delta h).
\end{equation}
By Assumption~A.1, $g$ is twice continuously differentiable. By the chain rule,
\begin{equation}
    g'(t) = \nabla Q(h + t \Delta h)\cdot \Delta h.
\end{equation}
Using the fundamental theorem of calculus,
\begin{equation}
\begin{aligned}
    Q(h + \Delta h) - Q(h)
    &= g(1) - g(0)
     = \int_0^1 g'(t)\,\mathrm{d}t \\
    &= \int_0^1 \nabla Q(h + t \Delta h)\cdot\Delta h\,\mathrm{d}t.
\end{aligned}
\label{eq:Q_diff_integral}
\end{equation}
Add and subtract $\nabla Q(h)$ inside the integrand:
\begin{equation}
\small
\begin{aligned}
Q(h+\Delta h)-Q(h)
&= \int_{0}^{1}
\!\bigl[\nabla Q(h)
  + (\nabla Q(h+t\Delta h)-\nabla Q(h))\bigr]\!\cdot\Delta h\,\mathrm{d}t \\
&= \nabla Q(h)\!\cdot\!\Delta h
 + \int_{0}^{1}\!(\nabla Q(h+t\Delta h)-\nabla Q(h))\!\cdot\!\Delta h\,\mathrm{d}t .
\end{aligned}
\label{eq:Q_diff_split}
\end{equation}

By the Cauchy--Schwarz inequality and Assumption~A.2,
\begin{equation}
\small
\begin{aligned}
\bigl|(\nabla Q(h+t\Delta h)-\nabla Q(h))\!\cdot\!\Delta h\bigr|
&\le \|\nabla Q(h+t\Delta h)-\nabla Q(h)\|\,\|\Delta h\| \\
&\le L_Q\, t\, \|\Delta h\|^2 .
\end{aligned}
\end{equation}

Hence,
\begin{equation}
\small
\int_{0}^{1} (\nabla Q(h+t\Delta h)-\nabla Q(h))\!\cdot\!\Delta h\,dt
\;\ge\;
- \int_{0}^{1} L_Q t\,\|\Delta h\|^{2}\,dt
= -\tfrac{L_Q}{2}\|\Delta h\|^{2}.
\end{equation}

\noindent Substituting into~\eqref{eq:Q_diff_split} yields~\eqref{eq:descent_lemma}, which completes the proof. \qedhere
\end{proof}

\subsubsection{Proof of Theorem 4.1}

We now prove the local, step-size--bounded positive correlation between confidence and reasoning quality.

\begin{theorem}[Local bounded positive gain]
\label{thm:local_positive_correlation}
Let Assumptions~A.1--A.3 hold at $h \in \mathcal{D}$. Define
\begin{equation}
    a := \nabla C(h)\cdot\nabla Q(h) > 0,
    \qquad
    b := \|\nabla C(h)\|^2 > 0.
    \label{eq:def_a_b}
\end{equation}
Then for any step size $\eta$ satisfying
\begin{equation}
    0 < \eta < \frac{2a}{L_Q b},
    \label{eq:eta_bound}
\end{equation}
the update~\eqref{eq:dmlr_update_conf} satisfies
\begin{equation}
    Q(h^{+}) > Q(h).
    \label{eq:Q_increase}
\end{equation}
In other words, within the bounded step-size interval~\eqref{eq:eta_bound}, a single step of confidence ascent strictly improves the reasoning-quality function $Q$.
\end{theorem}

\begin{proof}
Apply Lemma~\ref{lem:descent_lemma} with $\Delta h = \eta \nabla C(h)$, which gives
\begin{equation}
\begin{aligned}
    Q(h^{+}) - Q(h)
    &\ge 
    \nabla Q(h)\cdot (\eta \nabla C(h))
    - \frac{L_Q}{2}\eta^2 \|\nabla C(h)\|^2 \\
    &= \eta \,\nabla Q(h)\cdot\nabla C(h)
    - \frac{L_Q}{2}\eta^2 \|\nabla C(h)\|^2.
\end{aligned}
\label{eq:Q_diff_eta}
\end{equation}
Using the definitions in~\eqref{eq:def_a_b}, we obtain
\begin{equation}
    Q(h^{+}) - Q(h)
    \;\ge\;
    \eta a - \frac{L_Q}{2}\eta^2 b
    \;=:\;
    \phi(\eta).
    \label{eq:phi_eta}
\end{equation}
The function $\phi(\eta)$ is a concave quadratic in $\eta$ with roots at $\eta=0$ and
\begin{equation}
    \eta^\star = \frac{2a}{L_Q b}.
\end{equation}
Since $a>0$ and $b>0$, we have $\eta^\star>0$ and
\begin{equation}
    \phi(\eta) > 0
    \quad\text{for all } 0<\eta<\eta^\star.
\end{equation}
Therefore, for any $0<\eta<\eta^\star$, we obtain
\begin{equation}
    Q(h^{+}) - Q(h) \ge \phi(\eta) > 0,
\end{equation}
which proves~\eqref{eq:Q_increase}. 
\end{proof}

\subsubsection{Boundedness of the Positive-Correlation Region}

Theorem~\ref{thm:local_positive_correlation} shows that, at any point where the gradients of 
$C$ and $Q$ are positively aligned, there exists a \emph{bounded} step-size interval within which
ascending the confidence landscape strictly improves $Q$. We next argue that such a
positive-correlation property cannot hold globally in the presence of a ``confidently incorrect'' trap.

\begin{definition}[Confidently incorrect trap]
\label{def:trap}
A point $h_{\mathrm{trap}} \in \mathcal{D}$ is called a confidently incorrect trap if
\begin{align}
    &\nabla C(h_{\mathrm{trap}}) = 0,
    \quad
    \nabla^2 C(h_{\mathrm{trap}}) 
    \text{ is negative definite,}
    \label{eq:trap_conf_local_max} \\
    &Q(h_{\mathrm{trap}}) \ll \max_{h\in\mathcal{D}} Q(h).
    \label{eq:trap_low_quality}
\end{align}
In other words, $h_{\mathrm{trap}}$ is a strict local maximizer of the confidence landscape, but the corresponding reasoning quality is significantly lower than that of other regions.
\end{definition}

Consider the dynamical system induced by confidence ascent,
\begin{equation}
    h^{(t+1)} = h^{(t)} + \eta \nabla C\big(h^{(t)}\big),
    \label{eq:conf_dynamics}
\end{equation}
with a sufficiently small fixed step size $\eta>0$. Let $\mathcal{B}(h_{\mathrm{trap}})$
denote the basin of attraction of $h_{\mathrm{trap}}$ under~\eqref{eq:conf_dynamics}, i.e.,
the set of initial points $h^{(0)}$ for which $h^{(t)} \to h_{\mathrm{trap}}$ as $t\to\infty$.

\begin{proposition}[Positive correlation cannot hold everywhere in the basin]
\label{prop:bounded_region}
Suppose there exists a confidently incorrect trap $h_{\mathrm{trap}}$ in the sense of
Definition~\ref{def:trap}. Then it is impossible that
\begin{equation}
    \nabla C(h)\cdot\nabla Q(h) > 0,
    \quad \forall\, h \in \mathcal{B}(h_{\mathrm{trap}}).
    \label{eq:global_alignment_basin}
\end{equation}
Equivalently, there must exist points $h$ in the basin of attraction for which
\begin{equation}
    \nabla C(h)\cdot\nabla Q(h) \le 0.
\end{equation}
Thus, the region where confidence and reasoning-quality gradients are positively aligned
is necessarily bounded and cannot cover the entire basin of a confidently incorrect trap.
\end{proposition}

\begin{proof}
Assume, for the sake of contradiction, that~\eqref{eq:global_alignment_basin} holds, i.e.,
for all $h \in \mathcal{B}(h_{\mathrm{trap}})$ we have $\nabla C(h)\cdot\nabla Q(h) > 0$.
Pick an arbitrary $h^{(0)} \in \mathcal{B}(h_{\mathrm{trap}})$, and consider the trajectory
generated by~\eqref{eq:conf_dynamics}:
\begin{equation}
    h^{(t+1)} = h^{(t)} + \eta \nabla C(h^{(t)}).
\end{equation}

\noindent By Definition of $\mathcal{B}(h_{\mathrm{trap}})$, we have
\begin{equation}
    h^{(t)} \to h_{\mathrm{trap}}
    \quad\text{as } t\to\infty.
\end{equation}
Since $Q$ is continuous, it follows that
\begin{equation}
    \lim_{t\to\infty} Q(h^{(t)}) = Q(h_{\mathrm{trap}}).
    \label{eq:limit_Q_trap}
\end{equation}

\noindent On the other hand, for each $t$ such that $h^{(t)}\in\mathcal{D}$, Assumptions~A.1--A.2
hold at $h^{(t)}$ and the alignment condition
$\nabla C(h^{(t)})\cdot\nabla Q(h^{(t)})>0$ holds by~\eqref{eq:global_alignment_basin}. 
Hence, by Theorem~\ref{thm:local_positive_correlation}, there exists a step-size bound
$\eta_t^\star>0$ such that for all $0<\eta<\eta_t^\star$,
\begin{equation}
    Q(h^{(t+1)}) > Q(h^{(t)}).
\end{equation}
Taking $\eta$ small enough so that $\eta < \eta_t^\star$ for all $t$ along the trajectory,
we obtain a strictly increasing sequence:
\begin{equation}
    Q(h^{(0)}) < Q(h^{(1)}) < Q(h^{(2)}) < \cdots.
\end{equation}

Thus $\{Q(h^{(t)})\}_{t=0}^\infty$ is strictly increasing, and together with
\eqref{eq:limit_Q_trap}, we obtain $Q(h_{\mathrm{trap}})>Q(h^{(0)})$. 
This contradicts the low-quality condition \eqref{eq:trap_low_quality}, since
$h_{\mathrm{trap}}$ cannot be a confidently incorrect trap. 
Therefore, \eqref{eq:global_alignment_basin} must be false, and there exist
points $h\in\mathcal{B}(h_{\mathrm{trap}})$ such that
$\nabla C(h)\cdot\nabla Q(h)\le 0$. 
This completes the proof.
\end{proof}

\noindent\textbf{\textit{Proposition~\ref{prop:bounded_region} shows that the positive correlation between confidence
and reasoning quality, formalized via gradient alignment, is fundamentally a \emph{local}
property: it can hold within a bounded region (and within bounded step sizes) but cannot be
extended to the entire latent space in the presence of confidently incorrect traps.}}

\subsection{Visual Injection Enhances Confidence.}

Let $\mathcal{T}$ denote the latent reasoning states before visual injection,
$\hat{\mathcal{T}}$ the updated states after dynamic visual injection (DVI), and
$z_v$ the visual features.  
Let $Y$ be the target output random variable (e.g., the final answer).
We assume all variables are defined on a common probability space.

\noindent \textbf{Assumption B.1.}
The dynamic visual injection (DVI) module is constructed such that the
latent state after injection preserves at least as much visual information
as before:
\begin{equation}
    I(\hat{\mathcal{T}}; z_v) \;\ge\; I(\mathcal{T}; z_v).
    \label{eq:B1}
\end{equation}

\noindent \textbf{Assumption B.2.}
The visual features contain information about the target variable $Y$:
\begin{equation}
    I(z_v; Y) > 0.
    \label{eq:B2_visionY}
\end{equation}
Moreover, the task representation is trained so that the mutual information
between the latent state and $Y$ is a strictly increasing function of the
mutual information between the latent state and visual features.  
That is, there exists a strictly increasing function $g(\cdot)$ satisfying
\begin{equation}
    I(Y; \mathcal{T}) = g\!\big(I(\mathcal{T}; z_v)\big),\qquad
    I(Y; \hat{\mathcal{T}}) = g\!\big(I(\hat{\mathcal{T}}; z_v)\big).
    \label{eq:B2_g_function}
\end{equation}

\noindent \textbf{Assumption B.3.}
The model’s confidence objective is a strictly decreasing function of the
conditional entropy of $Y$ given the latent state:
\begin{equation}
    J_{\mathrm{conf}}(\mathcal{T}) 
    = \psi\!\big(H(Y\mid \mathcal{T})\big),\qquad
    J_{\mathrm{conf}}(\hat{\mathcal{T}})
    = \psi\!\big(H(Y\mid \hat{\mathcal{T}})\big),
    \label{eq:B3_psi}
\end{equation}
where $\psi(\cdot)$ is strictly decreasing.  
This reflects that lower predictive uncertainty corresponds to higher confidence.

\noindent We now show that~\eqref{eq:B1} implies
$J_{\mathrm{conf}}(\hat{\mathcal{T}})\ge J_{\mathrm{conf}}(\mathcal{T})$.

% ---------------- Lemma 1: MI(Z,T) -> MI(Y,T) ----------------
\begin{lemma}[From vision MI to task MI]
\label{lem:MI_to_task}
Under Assumption~B.2, if
$I(\hat{\mathcal{T}}; z_v)\ge I(\mathcal{T}; z_v)$, then
\begin{equation}
    I(Y;\hat{\mathcal{T}})\;\ge\; I(Y;\mathcal{T}).
    \label{eq:task_MI_increase}
\end{equation}
\end{lemma}

\begin{proof}
By~\eqref{eq:B2_g_function},
\[
I(Y;\hat{\mathcal{T}}) = g\!\big(I(\hat{\mathcal{T}}; z_v)\big),
\quad
I(Y;\mathcal{T}) = g\!\big(I(\mathcal{T}; z_v)\big).
\]
Since $g$ is strictly increasing and
$I(\hat{\mathcal{T}}; z_v)\ge I(\mathcal{T}; z_v)$, we obtain~\eqref{eq:task_MI_increase}.
\end{proof}

% ---------------- Lemma 2: task MI -> conditional entropy ----------------
\begin{lemma}[Task MI and conditional entropy]
\label{lem:MI_entropy}
If $I(Y;\hat{\mathcal{T}})\ge I(Y;\mathcal{T})$, then
\begin{equation}
    H(Y\mid \hat{\mathcal{T}}) \;\le\; H(Y\mid \mathcal{T}).
    \label{eq:entropy_decrease}
\end{equation}
\end{lemma}

\begin{proof}
By the identity $I(Y;U) = H(Y) - H(Y\mid U)$ for any random variable $U$,
\[
H(Y\mid U) = H(Y) - I(Y;U).
\]
Thus
\[
H(Y\mid \hat{\mathcal{T}}) = H(Y) - I(Y;\hat{\mathcal{T}}),
\quad
H(Y\mid \mathcal{T}) = H(Y) - I(Y;\mathcal{T}).
\]
If $I(Y;\hat{\mathcal{T}})\ge I(Y;\mathcal{T})$, then
$H(Y\mid \hat{\mathcal{T}})\le H(Y\mid \mathcal{T})$, which is~\eqref{eq:entropy_decrease}.
\end{proof}

% ---------------- Lemma 3: conditional entropy -> confidence ----------------
\begin{lemma}[Conditional entropy and confidence]
\label{lem:entropy_confidence}
Under Assumption~B.3, if 
$H(Y\mid \hat{\mathcal{T}})\le H(Y\mid \mathcal{T})$, then
\begin{equation}
    J_{\mathrm{conf}}(\hat{\mathcal{T}})
    \;\ge\;
    J_{\mathrm{conf}}(\mathcal{T}).
    \label{eq:conf_increase}
\end{equation}
\end{lemma}

\begin{proof}
By~\eqref{eq:B3_psi},
\[
J_{\mathrm{conf}}(\hat{\mathcal{T}})
= \psi\!\big(H(Y\mid \hat{\mathcal{T}})\big),
\quad
J_{\mathrm{conf}}(\mathcal{T})
= \psi\!\big(H(Y\mid \mathcal{T})\big).
\]
Since $\psi$ is strictly decreasing and
$H(Y\mid \hat{\mathcal{T}})\le H(Y\mid \mathcal{T})$,
we obtain~\eqref{eq:conf_increase}.
\end{proof}

% ---------------- Final theorem ----------------
\begin{theorem}[For Theorem 4.2]
\label{thm:theorem42_formal}
Under Assumptions~B.1--B.3, if
\begin{equation}
    I(\hat{\mathcal{T}}; z_v)\;\ge\; I(\mathcal{T}; z_v),
\end{equation}
then
\begin{equation}
    J_{\mathrm{conf}}(\hat{\mathcal{T}})
    \;\ge\;
    J_{\mathrm{conf}}(\mathcal{T}).
\end{equation}
\end{theorem}

\begin{proof}
From Assumption~B.1 and Lemma~\ref{lem:MI_to_task},
$I(\hat{\mathcal{T}}; z_v)\ge I(\mathcal{T}; z_v)$
implies $I(Y;\hat{\mathcal{T}})\ge I(Y;\mathcal{T})$.
By Lemma~\ref{lem:MI_entropy}, this yields
$H(Y\mid \hat{\mathcal{T}})\le H(Y\mid \mathcal{T})$.
Finally, Lemma~\ref{lem:entropy_confidence} implies
$J_{\mathrm{conf}}(\hat{\mathcal{T}})\ge J_{\mathrm{conf}}(\mathcal{T})$.
\end{proof}

\noindent Theorem~\ref{thm:theorem42_formal} formalizes that, under mild assumptions,
DVI-induced increases in the mutual information between latent states and visual
features translate into lower predictive uncertainty about the target output,
and thus into higher expected confidence.

\subsection{Derivation of the Latent Policy Gradient and Update Rule}
\label{app:latent_pg_full}
We give a detailed derivation of the gradient used to update the latent
thought vectors $H$ (e.g., latent think tokens) via a Gaussian policy gradient method.
\vspace{-0.2cm}
\paragraph{Objective.}
Let $A$ denote the sampled latent action (a perturbed latent thought
vector) and $R(A)$ the corresponding reward or confidence score. Given a
stochastic policy $\pi(A\mid H)$ over actions $A$ conditioned on the
current latent vector $H$, we define the expected objective
\begin{equation}
    J(H)
    := \mathbb{E}_{A\sim\pi(\cdot\mid H)}[R(A)].
    \label{eq:latent_obj}
\end{equation}

Our goal is to derive $\nabla_H J(H)$ and obtain a practical estimator
for use in gradient ascent.

%---------------------------------------------------------------
\subsubsection{Policy Gradient Form}
\begin{lemma}[Policy gradient identity]
\label{lem:pg_identity_H}
Under mild regularity assumptions (allowing differentiation under the
integral sign), the gradient of $J(H)$ in~\eqref{eq:latent_obj} satisfies
\begin{equation}
    \nabla_H J(H)
    = \mathbb{E}_{A\sim\pi(\cdot\mid H)}
      \big[ R(A)\,\nabla_H \log \pi(A\mid H) \big].
    \label{eq:pg_theorem_H}
\end{equation}
\end{lemma}

\begin{proof}
By definition of expectation,
\begin{align}
    J(H)
    &= \int R(A)\,\pi(A\mid H)\,dA.
\end{align}
Differentiating w.r.t.\ $H$ yields
\begin{align}
    \nabla_H J(H)
    &= \int R(A)\,\nabla_H \pi(A\mid H)\,dA.
\end{align}
Using the log-derivative identity
$\nabla_H \pi(A\mid H) = \pi(A\mid H)\,\nabla_H \log\pi(A\mid H)$, we obtain
\begin{align}
    \nabla_H J(H)
    &= \int R(A)\,\pi(A\mid H)\,\nabla_H \log\pi(A\mid H)\,dA \\
    &= \mathbb{E}_{A\sim\pi(\cdot\mid H)}
       \big[ R(A)\,\nabla_H \log \pi(A\mid H) \big],
\end{align}
which is~\eqref{eq:pg_theorem_H}.
\end{proof}

\subsubsection{Gaussian Latent Policy}
We now instantiate the policy as an isotropic Gaussian perturbation of
$H$. Specifically, we sample
\begin{equation}
    A = H + \xi^{(t)},
    \qquad
    \xi^{(t)} \sim \mathcal{N}(0,\sigma^{2}I),
    \label{eq:gaussian_policy_A}
\end{equation}
which is equivalent to
\begin{equation}
    \pi(A\mid H)
    = \mathcal{N}(A;\,H,\sigma^{2}I).
\end{equation}

\noindent We next compute the gradient of the log-probability under this Gaussian
policy.

\begin{lemma}[Gradient of Gaussian log-probability]
\label{lem:gaussian_score}
For the Gaussian policy in~\eqref{eq:gaussian_policy_A}, the
log-probability gradient satisfies
\begin{equation}
    \nabla_H \log \pi(A\mid H)
    = \frac{A - H}{\sigma^{2}}
    = \frac{\xi}{\sigma^{2}}.
    \label{eq:gaussian_score}
\end{equation}
\end{lemma}

\begin{proof}
The log-density of $\pi(A\mid H)$ is
\begin{equation}
    \log\pi(A\mid H)
    = -\frac{1}{2\sigma^{2}}\|A-H\|^{2} + \mathrm{const},
\end{equation}
where the constant does not depend on $H$. Differentiating w.r.t.\ $H$,
\begin{align}
    \nabla_H \log\pi(A\mid H)
    &= -\frac{1}{2\sigma^{2}}
       \nabla_H \|A-H\|^{2} \\
    &= -\frac{1}{2\sigma^{2}}
       \nabla_H (A-H)^\top(A-H) \\
    &= -\frac{1}{2\sigma^{2}}\cdot(-2)(A-H) \\
    &= \frac{A-H}{\sigma^{2}}.
\end{align}
Using $A = H + \xi$ from~\eqref{eq:gaussian_policy_A} yields
$A-H = \xi$, giving~\eqref{eq:gaussian_score}.
\end{proof}

\noindent  Substituting~\eqref{eq:gaussian_score} into the policy gradient
expression~\eqref{eq:pg_theorem_H}, we obtain
\begin{align}
    \nabla_H J(H)
    &= \mathbb{E}_{A\sim\pi(\cdot\mid H)}
       \big[ R(A)\,\nabla_H \log\pi(A\mid H) \big] \\
    &= \mathbb{E}_{\epsilon\sim\mathcal{N}(0,\sigma^{2}I)}
       \left[
           R(H+\xi)\,\frac{\xi}{\sigma^{2}}
       \right].
    \label{eq:latent_pg_gaussian}
\end{align}

\subsubsection{Monte Carlo Estimator from a Single Sample}

In practice, the expectation in~\eqref{eq:latent_pg_gaussian} is
approximated via Monte Carlo sampling. At optimization step $t$, we draw
a single Gaussian noise sample
\[
\mathcal{\xi}^{(t)}\sim \mathcal{N}(0,\sigma^{2}I),
\qquad
A^{(t)} = H^{(t)} + \xi^{(t)},
\]
and construct the one-sample estimator
\begin{equation}
    \nabla_H J(H^{(t)})
    \approx R\big(H^{(t)} + \xi^{(t)}\big)
             \frac{\xi^{(t)}}{\sigma^{2}}.
    \label{eq:latent_pg_MC}
\end{equation}

\begin{lemma}[Unbiasedness of the Monte Carlo estimator]
\label{lem:unbiased_MC}
Let $\mathcal{\xi}^{(t)}\sim \mathcal{N}(0,\sigma^{2}I)$ and define the
estimator
\(
    \widehat{g}^{(t)}
    = R(H^{(t)} + \xi^{(t)})
      \xi^{(t)}/\sigma^{2}.
\)
Then
\begin{equation}
    \mathbb{E}[\widehat{g}^{(t)}]
    = \nabla_H J(H^{(t)}),
\end{equation}
where the expectation is taken over $\xi^{(t)}$.
\end{lemma}

\begin{proof}
By definition and linearity of expectation,
\begin{align}
    \mathbb{E}[\widehat{g}^{(t)}]
    &= \mathbb{E}_{\xi^{(t)}}
       \left[
           R(H^{(t)} + \xi^{(t)})
           \frac{\epsilon^{(t)}}{\sigma^{2}}
       \right]
     = \nabla_H J(H^{(t)}),
\end{align}
where the last equality is exactly
\eqref{eq:latent_pg_gaussian} evaluated at $H = H^{(t)}$.
\end{proof}

\subsubsection{Gradient Ascent Update}

Finally, we perform standard gradient ascent on the latent vector $H$.
With learning rate $\eta>0$, the update at step $t$ is
\begin{equation}
    H^{(t+1)}
    = H^{(t)}
      + \eta \cdot
        R\big(H^{(t)} + \xi^{(t)}\big)
        \frac{\xi^{(t)}}{\sigma^{2}}.
    \label{eq:latent_update_rule}
\end{equation}

\noindent This update moves the latent thought vector $H$ in the direction of
noisy perturbations $\xi^{(t)}$ that lead to higher reward, as
estimated by the Monte Carlo policy gradient.

\section{More detail about Visual Injection.}
\label{e}

\textbf{Different Strategies Discussion.} We explore two visual injection mechanisms to identify which strategy best supports multi-step latent reasoning optimization. \textbf{\textit{The first strategy}}, ultimately adopted in our DVI Strategy is a dynamic visual patch search-and-refinement procedure organized around the three-phase cycle shown in Algorithm~\ref{latent_reasoning}. At the start of the reasoning trajectory, we use the initial latent think token’s cross-modal attention to select the top-$m$ visually relevant patches, forming an initial candidate set $\mathcal{Z}_{\text{cand}}$ and an initial best-patch set $\mathcal{V}_{\text{best}}$. This initialization provides a reasonable but unconverged visual hypothesis, ensuring that the search begins from semantically meaningful regions while preserving room for subsequent refinement.

During each latent-update iteration, we recompute attention based on the updated hidden state $\mathcal{T}^{(t)}$ and resample a fresh batch of candidate patches $\mathcal{Z}_{\text{cand}} = \{Z_1, \dots, Z_m\}$. This dynamic resampling allows the visual hypothesis to adapt in tandem with the evolving reasoning context, preventing the model from becoming locked to the potentially imperfect initial attention pattern. We then construct an expanded latent representation $\tilde{\mathcal{T}}^{(t)}_l = [\mathcal{T}^{(t)}_l, \mathcal{Z}_{\text{cand}}, \mathcal{V}_{\text{best}}]$ and compute its reward $r = \mathcal{R}(\mathcal{Z}, \mathcal{Q}, \tilde{\mathcal{T}}^{(t)}_l)$, which acts as a task-level criterion for validating whether the newly sampled patches truly enhance reasoning performance.

If the obtained reward surpasses the historical best value $r_{\text{best}}$, we incorporate the new patches into the best-patch set $\mathcal{V}_{\text{best}}$ and carry forward the updated latent representation into the next iteration; otherwise, we revert the update to preserve previously verified visual evidence. Over multiple iterations, $\mathcal{V}_{\text{best}}$ is progressively refined under the interplay of exploration (resampling new patches) and exploitation (reward-based filtering), ultimately converging to a compact and semantically stable set of visual anchors. This dynamic best-patch optimization provides consistent and reward-validated visual grounding throughout the reasoning chain, producing far more reliable behavior than strategies that rely solely on per-step attention signals.

\textbf{\textit{The second strategy}} adopts a more direct formulation: at each iteration, it simply injects the visual patch that receives the highest attention, thereby emulating the attention-guided visual grounding commonly used in existing multimodal models. These two strategies thus represent fundamentally different paradigms for visual injection—one progressively optimizes visual evidence through reward-driven refinement, whereas the other relies on instantaneous attention-based selection. \textit{Given their distinct underlying principles, it is necessary to systematically compare their behaviors and performance characteristics.}

\noindent \textbf{Results Analysis.} As shwon in Table~\ref{tab:stability_compare} and \ref{tab:perf_compare}, we observe that injecting the highest attention visual patch at each iteration appears straightforward but suffers from fundamental instability. Attention scores are highly local and easily influenced by visual saliency, token co-adaptation, and early stage noise, causing the most attended region to deviate from the truly task relevant area. Because this strategy depends entirely on instantaneous attention distributions, the injected visual evidence fluctuates across iterations, resulting in drifting visual focus and compounding errors when early attention is misaligned. In contrast, our dynamic visual injection strategy, based on iterative patch resampling, reward evaluation, and best patch refinement, provides a far more stable grounding mechanism. By validating new patches through task level reward rather than noisy attention signals, the model progressively converges to a semantically consistent and task relevant set of visual anchors. This search, verify, refine process stabilizes visual grounding across iterations, prevents error amplification, and yields markedly more reliable reasoning trajectories. Empirically, the dynamic best patch optimization consistently outperforms attention based injection, demonstrating superior robustness and significantly higher multimodal reasoning accuracy.

\begin{table}[h]
\centering
\caption{Performance comparison between the two visual injection strategies on the MMVP Benchmark.}
\vspace{0.1cm}
\resizebox{0.8\linewidth}{!}{
\begin{tabular}{l c}
\toprule
\textbf{Method} & \textbf{Accuracy $\uparrow$} \\
\midrule
Attention-based  & 66.3 \\
Dynamic best-patch (ours) & \textbf{74.8} \\
\bottomrule
\end{tabular}}
\label{tab:perf_compare}
\end{table}

\begin{table}[h]
\centering
\caption{Stability metrics across iterative reasoning on the MathVision Benchmark. Lower is better.}
\vspace{0.1cm}
\resizebox{\linewidth}{!}{
\begin{tabular}{l cc}
\toprule
\textbf{Method} & \textbf{Attention Variance $\downarrow$} & \textbf{Drift Distance $\downarrow$} \\
\midrule
Attention-based  & 0.48 & 12.3 \\
Dynamic best-patch (ours) & \textbf{0.12} & \textbf{2.1} \\
\bottomrule
\end{tabular}}
\label{tab:stability_compare}
\end{table}

\noindent \textbf{In-depth Analysis.} When designing DVI, a seemingly reasonable question arises: since visual injection is intended to supply additional evidence when the model becomes uncertain, can confidence be used as a trigger signal so that visual patch optimization is performed only when confidence falls below a predefined threshold? This confidence-triggered variant appears more economical and aligns with the intuition of “inject vision only when needed.” However, our experiments (as shown in Table~\ref{tab:confidence_vs_continuous}) reveal that this strategy leads to noticeably worse convergence behavior. The optimization of visual patches in DVI is inherently continuous and incremental, relying on multi-step accumulation rather than sporadic corrections. Restricting updates to low confidence moments breaks this continuity. In practice, the model often enters a suboptimal regime in which confidence remains superficially stable while the underlying visual grounding is still misaligned, delaying or even preventing necessary corrections. As a result, the overall number of iterations required for convergence increases, and the final visual anchor becomes less stable and less semantically consistent than under the continuous best-patch refinement strategy. These findings indicate that confidence is not a reliable indicator for when visual refinement should occur, and maintaining per-iteration optimization is essential for achieving robust, efficient, and semantically coherent convergence in DVI.

\begin{table}[t]
\centering
\caption{
Comparison between confidence-triggered refinement and continuous DVI optimization on the MMVP Benchmark.
\textbf{Iter.} refers to the number of iterations required for convergence,
\textbf{Cons.} refers to the cross-iteration patch consistency,
\textbf{Err.} refers to the grounding error,
and \textbf{Acc.} refers to the final answer accuracy.
Continuous optimization converges faster, produces more stable visual anchors, and achieves higher accuracy.
}
\vspace{0.15cm}
\resizebox{\linewidth}{!}{
\begin{tabular}{lcccc}
\toprule
\textbf{Method} &
\textbf{Iter. $\downarrow$} &
\textbf{Cons. $\uparrow$} &
\textbf{Err. $\downarrow$} &
\textbf{Acc. $\uparrow$} \\
\midrule
Confidence-triggered update & 14.2 & 0.46 & 0.27 & 68.1 \\
Continuous best-patch (ours) & \textbf{9.3} & \textbf{0.83} & \textbf{0.12} & \textbf{74.8} \\
\bottomrule
\end{tabular}}
\label{tab:confidence_vs_continuous}
\vspace{-0.2cm}
\end{table}

\begin{algorithm}[h] 
\caption{\small \textbf{\textit{Dynamic Multimodal Latent Reasoning}}}
\small
\label{latent_reasoning}
\DontPrintSemicolon
\SetAlgoVlined
\SetKwProg{Fn}{}{}{}
\KwRequire{Image embeddings $\mathcal{Z}$, text embeddings $\mathcal{Q}$, 
latent tokens $\mathcal{T}_l$, learning rate $\eta$, 
iterations $T$, best visual patch $\mathcal{V}_{\textit{best}}$, 
top-$k$ probability $\mathrm{Top}_k(\mathcal{P}_i)$, the number of candidate patches 
$m$}

\vspace{2pt}
$\mathrm{Top}_k(\mathcal{P}_i) = \pi_{\theta}([\mathcal{Q}, \mathcal{Z}, \mathcal{T}])$; 
$r \gets \mathcal{R}(\mathcal{P}_{i})$ 
\hspace{1.em}\textcolor{blue}{$\triangleright$ reward}

\vspace{3pt}
\textcolor{darkgray}{\# \textit{Latent Policy Gradient Optimization}}\;

\For{$T = 1 \dots t$}{
    $\epsilon \sim \mathcal{N}(0,\sigma^2 I)$ 
    \hfill \textcolor{blue}{$\triangleright$ latent perturbation}

    $\mathcal{T}^{(t)\prime} \gets \mathcal{T}^{(t)} + \epsilon$

    $\mathcal{T}^{(t)} \gets \mathcal{T}^{(t)} + 
    \eta \nabla_{\mathcal{T}^{(t)}} \mathcal{J}(\mathcal{T}^{(t)})$
    \hfill \textcolor{blue}{$\triangleright$ latent update}

    \vspace{2pt}
    \textcolor{darkgray}{\# \textit{Dynamic Visual Injection}}\;
    
    $\mathcal{V}_{\textit{best}} \gets \text{Initialize}(\mathcal{T}^{(0)}, m)$ \hfill \textcolor{blue}{$\triangleright$ initialize best patch}

    \For{$L = 1 \dots l$}{
    
       $\mathcal{Z}_{\textit{cand}} \gets 
        \text{AttentionSelect}(\mathcal{T}_l^{(t)}, m)$
        \hfill \textcolor{blue}{$\triangleright$ select $m$ candidate visual patches}

        $\tilde{\mathcal{T}_l}^{(t)} \gets [\mathcal{T}_l^{(t)}, \mathcal{Z}_{\textit{cand}}, \mathcal{V}_{\textit{best}}]$

        $r \gets \mathcal{R}(\mathcal{Q}, \mathcal{Z}, \tilde{\mathcal{T}^{(t)}_l})$

        \If{$r > r_{\text{best}}$}{
            $\mathcal{V}_{\textit{best}} \gets \mathcal{V}_{\textit{best}} \cup \mathcal{Z}_{\textit{cand}}$;

            $\mathcal{T}_l^{(t)} \gets \tilde{\mathcal{T}_l}^{(t)}$
            \hfill \textcolor{blue}{$\triangleright$ update best}
        }
        \Else{
            $\mathcal{T}_l^{(t)} \gets 
            [\mathcal{T}_l^{(t)}, \mathcal{V}_{\textit{best}}]$
            \hfill \textcolor{blue}{$\triangleright$ revert to previous best}
        }
        \vspace{1pt}
    }
    \vspace{2pt}
}

\vspace{3pt}
$\mathcal{X} \gets \text{Decode}(\mathcal{T}^{(t)}, \mathcal{Z}, \mathcal{Q})$

\Return $\mathcal{X}$
\end{algorithm}

\section{Case Study}
\label{f}
In this section, we provide qualitative examples to illustrate the behavior of the DMLR-enhanced model across different task settings. As shown in \textbf{Example 1-8}, the model demonstrates more reliable and fine-grained visual perception, supported by stable and semantically coherent internal representations. In reasoning scenarios, the model produces accurate and well-structured logical steps. Importantly, the dynamic latent reasoning mechanism suppresses unnecessary expansions, resulting in concise and task-focused reasoning chains rather than overly long or speculative explanations. Taken together, these examples highlight that DMLR simultaneously strengthens perceptual quality and reasoning reliability, while ensuring that the generated reasoning remains efficient and free from redundant elaboration.

\begin{figure*}[t]
    \centering
    \begin{mdframed}[
        frametitle={Perception Example 1},
        frametitlebackgroundcolor=lightyellow,
        backgroundcolor=gray!10,
        linecolor=gray,
        frametitlerule=true,
        frametitlefont=\color{black}\bfseries
    ]
        \centering
        \small
        % 图片
        \includegraphics[width=0.4\linewidth]{img/example/mmvp1.jpg} 
        \captionof{figure}{\centering Question: Is the door of the truck cab open?\\Options: (a) Yes (b) No\\Ground Truth: (a)       \label{fig1}} 
  
        \flushleft

    \noindent\textbf{Qwen2.5-VL-7B-Instruct + DMLR:}

    The image shows a garbage truck with its door on the side of the cab open. This is evident from the visible interior and the fact that the door is not aligned with the rest of the vehicle's structure, indicating it has been opened. The answer is \boxed{a}. 

    \vspace{1em}

    \noindent\textbf{Qwen2.5-VL-7B-Instruct:}

    To determine if the door of the truck cab is open, let's analyze the image step by step:
    \begin{enumerate}
        \item Observe the side of the truck where the door would be located.
        \item Look for any visible gap or opening that would indicate the door is not closed.
        \item Check the overall appearance of the door; if it is open, there should be a noticeable gap or difference in the alignment compared to when it is closed.
    \end{enumerate}

    Upon examining the image:
    \begin{itemize}
        \item The door appears to be aligned with the rest of the truck's body.
        \item There is no visible gap or difference in the alignment that would suggest the door is open.
    \end{itemize}

    Based on this analysis, the door of the truck cab does not appear to be open.

    \noindent\texttt{<think>} The door of the truck cab is not open.\texttt{</think>}
    \noindent\texttt{<answer>} No\texttt{</answer>}

    \noindent\boxed{\text{No}}
       \label{fig1}
    \end{mdframed}
\end{figure*}

\begin{figure*}[t]
    \centering
    \begin{mdframed}[
        frametitle={Perception Example 2},
        frametitlebackgroundcolor=lightyellow,
        backgroundcolor=gray!10,
        linecolor=gray,
        frametitlerule=true,
        frametitlefont=\color{black}\bfseries
    ]
        \centering
        \small
        % 1. 替换为指定的图片路径
        \includegraphics[width=0.4\linewidth]{img/example/mmvp2.jpg} 
        \captionof{figure}{\centering Question: What is the woman more likely to do?\\Options: (a) Bump (b) Set\\Ground Truth: (b)} 
        \flushleft

    \noindent\textbf{Qwen2.5-VL-7B-Instruct + DMLR:}

    In the image, a volleyball game is in progress. The woman in focus appears to be positioned near the net, likely preparing to receive the ball. In volleyball, there are two primary actions for receiving the ball: bumping and setting.

    \begin{itemize}
        \item Bumping involves using the hands to cushion the ball, typically when it's coming down from above.
        \item Setting involves using the hands to lift the ball upward, usually when it's coming down or is at a low trajectory.
    \end{itemize}

    Given her position and the context of the game, she seems to be preparing to set the ball rather than bump it. This is because setting is a more strategic action that allows the team to control the direction and height of the ball, which is crucial for passing and spiking effectively.

    Therefore, based on the context and typical volleyball gameplay, the woman is more likely to set the ball. The answer is \boxed{b) \text{Set}}.

    \vspace{1em}

    \noindent\textbf{Qwen2.5-VL-7B-Instruct:}

    To determine what the woman is more likely to do, let's analyze the image step by step:
    \begin{enumerate}
        \item \textbf{Identify the context}: The image shows a volleyball game in progress. The woman in the foreground appears to be preparing to receive the ball, which is above her head.
        \item \textbf{Understand the volleyball terms}:
        \begin{itemize}
            \item \textbf{Bumping}: This is when a player uses their hands to pass the ball back into play after it has been spiked or set.
            \item \textbf{Setting}: This is when a player uses their hands to pass the ball high into the air for a teammate to spike.
        \end{itemize}
        \item \textbf{Analyze the woman's position}:
        \begin{itemize}
            \item The ball is above her head, and she appears to be in a position where she is about to use her hands to pass the ball.
            \item Her body posture suggests she is preparing to use her hands to control the ball, which is characteristic of bumping rather than setting.
        \end{itemize}
        \item \textbf{Consider the options}:
        \begin{itemize}
            \item Option (a) Bump: This aligns with her posture and the context of the game.
            \item Option (b) Set: This would require her to position herself differently and use her hands in a way that is not consistent with her current stance.
        \end{itemize}
    \end{enumerate}

    Given the analysis, the woman is more likely to bump the ball.

    \noindent\texttt{<think>} The woman is in a position to use her hands to control the ball, which is characteristic of bumping rather than setting.\texttt{</think>}
    \noindent\texttt{<answer>} \boxed{a} \texttt{</answer>}

    \end{mdframed}
\end{figure*}

\begin{figure*}[t]
    \centering
    \begin{mdframed}[
        frametitle={Perception Example 3},
        frametitlebackgroundcolor=lightyellow,
        backgroundcolor=gray!10,
        linecolor=gray,
        frametitlerule=true,
        frametitlefont=\color{black}\bfseries
    ]
        \centering
        \small
        \includegraphics[width=0.4\linewidth]{img/example/mmvp3.jpg} 
        \captionof{figure}{\centering Question: Can you see the caudal fin of the shark in the image?\\Options: (a) Yes (b) No\\Ground Truth: (a) Yes} 
        \flushleft

    % --- DMLR 模型部分 ---
    \noindent\textbf{VLAA-Thinker-Qwen2.5VL-7B + DMLR:}
    To determine if the caudal fin of the shark is visible in the image, I need to analyze the orientation and positioning of the shark. The caudal fin is typically located at the rear end of the shark's body. In this image, the shark is shown from a side angle, and its tail is clearly visible extending towards the bottom right corner of the image. This indicates that the caudal fin can be seen.

    Therefore, based on the visibility of the tail in the image, the answer is (a) Yes. The answer is \boxed{a}.

    \vspace{1em}

    % --- Base 模型部分 ---
    \noindent\textbf{VLAA-Thinker-Qwen2.5VL-7B:}

    \texttt{<think>}
    To determine if the caudal fin of the shark is visible in the image, I need to analyze the orientation and position of the shark. The caudal fin is typically located at the rear end of the shark's body. In this image, the shark is shown from a side angle, and the caudal fin is not directly visible due to the angle and the positioning of the shark's body. However, the tail area, which includes the caudal fin, can be inferred to be present based on the overall shape and structure of the shark.

    Given the options:
    (a) Yes
    (b) No

    The caudal fin is not directly visible in the image, but it can be inferred to be present based on the shark's anatomy and the angle of the image.

    Therefore, the correct answer is (b) No.
    \texttt{</think>}

    \noindent\texttt{<answer>} (b) No \texttt{</answer>}

    \end{mdframed}
\end{figure*}

\begin{figure*}[t]
    \centering
    \begin{mdframed}[
        frametitle={Perception Example 4},
        frametitlebackgroundcolor=lightyellow,
        backgroundcolor=gray!10,
        linecolor=gray,
        frametitlerule=true,
        frametitlefont=\color{black}\bfseries
    ]
        \centering
        \small
        \includegraphics[width=0.4\linewidth]{img/example/mmvp4.jpg} 
        \captionof{figure}{\centering Question: Is the flame of the match more round or thin?\\Options: (a) Round (b) Thin\\Ground Truth: (b) Thin} 
        \flushleft

    % --- DMLR 模型部分 ---
    \noindent\textbf{VLAA-Thinker-Qwen2.5VL-7B + DMLR:}

    To determine whether the flame of the match is more round or thin, let's analyze the image. The flame appears elongated and narrow at the top, which suggests it is not round but rather thin in width. The base of the flame seems wider, but the overall shape of the flame is more elongated than circular.

    Therefore, based on the visual analysis, the flame of the match is more thin. The answer is \boxed{b}.

    \vspace{1em}

    % --- Base 模型部分 ---
    \noindent\textbf{VLAA-Thinker-Qwen2.5VL-7B:}

    \texttt{<think>}
To determine whether the flame of the match is more round or thin, let's analyze the image. The flame appears to have a distinct shape where the base is wider and narrows towards the top, which is characteristic of a conical shape. This shape suggests that the flame is not uniformly round but rather has a tapering effect from bottom to top. Given this observation, the flame can be described as more round at its base and becoming thinner towards the top, but overall, it maintains a rounded appearance.

Therefore, based on the visual analysis, the flame of the match is more round.
    \texttt{</think>}

    \noindent\texttt{<answer>} \boxed{a} \texttt{</answer>}

    \end{mdframed}
\end{figure*}

\begin{figure*}[t]
    \centering
    \begin{mdframed}[
        frametitle={Reasoning Example 1},
        frametitlebackgroundcolor=lightyellow,
        backgroundcolor=gray!10,
        linecolor=gray,
        frametitlerule=true,
        frametitlefont=\color{black}\bfseries
    ]
        \centering
        \small
        \includegraphics[width=0.4\linewidth]{img/example/mathvision1.png} 
        \captionof{figure}{\centering Question: A beetle walks along the edges of a cube. Starting from point $P$ it first moves in the direction shown. At the end of each edge it changes the direction in which it turns, turning first right then left, then right etc. Along how many edges will it walk before it returns to point $P$?\\Ground Truth: 6} 
        \flushleft

    \noindent\textbf{Qwen2.5-VL-7B-Instruct + DMLR:}

    The beetle starts at point P and moves along the edges of a cube, changing direction according to the pattern: right, left, right. To determine how many edges it will walk before returning to point P, we need to trace its path.

    \begin{enumerate}
        \item The beetle starts at P.
        \item It moves right to an adjacent vertex.
        \item It turns left to another adjacent vertex.
        \item It turns right again to reach a third adjacent vertex.
        \item It continues this pattern until it returns to P.
    \end{enumerate}

    By examining the cube's structure, we can see that after 6 edges, the beetle will return to point P because each face of the cube has 4 edges, and the beetle covers all edges of one face before repeating the pattern on the next face.

    Thus, the beetle will walk along 6 edges before returning to point P.

    \vspace{1em}

    \noindent\textbf{Qwen2.5-VL-7B-Instruct:}

    To solve this problem, we need to track the beetle's path as it moves along the edges of the cube, changing its direction according to the specified pattern (right, left, right, etc.).

    \begin{enumerate}
        \item \textbf{Initial Position and Direction}: The beetle starts at point $P$ and moves in the direction indicated by the arrow, which is along one edge of the cube.
        \item \textbf{First Edge}: The beetle moves along the first edge and changes direction to the right.
        \item \textbf{Second Edge}: Moving right, the beetle now faces a new direction and continues along the next edge.
        \item \textbf{Third Edge}: After moving along the second edge, the beetle changes direction again to the left.
        \item \textbf{Fourth Edge}: Moving left, the beetle now faces another new direction and continues along the next edge.
        \item \textbf{Fifth Edge}: After moving along the third edge, the beetle changes direction to the right again.
        \item \textbf{Sixth Edge}: Moving right, the beetle now faces yet another new direction and continues along the next edge.
        \item \textbf{Seventh Edge}: After moving along the fourth edge, the beetle changes direction to the left once more.
        \item \textbf{Eighth Edge}: Moving left, the beetle now faces a new direction and continues along the next edge.
    \end{enumerate}

    At this point, the beetle has completed a full cycle around the cube, returning to point $P$.

    The beetle will have walked along 8 edges before it returns to point $P$. The answer is 8.

    \end{mdframed}
\end{figure*}

\begin{figure*}[p]
    \centering
    \begin{mdframed}[
        frametitle={Reasoning Example 2},
        frametitlebackgroundcolor=lightyellow,
        backgroundcolor=gray!10,
        linecolor=gray,
        frametitlerule=true,
        frametitlefont=\color{black}\bfseries
    ]
        \centering
        \small
        \includegraphics[width=0.4\linewidth]{img/example/mathvision2.png} 
        \captionof{figure}{\centering Question: In the adjoining plane figure, sides $AF$ and $CD$ are parallel, as are sides $AB$ and $EF$, and sides $BC$ and $ED$. Each side has length of 1. Also, $\measuredangle FAB = \measuredangle BCD = 60^\circ$. The area of the figure is\\Options: A. $\frac{\sqrt{3}}{2}$, B. $1$, C. $\frac{3}{2}$, D. $\sqrt{3}$, E. $2$\\Ground Truth: D. $\sqrt{3}$} 
        \flushleft

    \noindent\textbf{Qwen2.5-VL-7B-Instruct + DMLR:}

    The given figure is a hexagon with sides of equal length and specific angles. To find the area, we can break it down into simpler shapes like triangles and parallelograms.

    \begin{enumerate}
        \item \textbf{Identify the shape}: The hexagon can be divided into two congruent trapezoids by drawing a diagonal from one vertex to the opposite non-adjacent vertex.
        \item \textbf{Calculate the area of each trapezoid}:
        \begin{itemize}
            \item Each trapezoid has parallel sides of 1 unit (AF and CD), and the height can be calculated using the properties of the 60° angle.
            \item The height of the trapezoid can be found using the sine of 60° in a right triangle formed by dropping a perpendicular from one of the vertices to the base.
            \item The area of one trapezoid is $\frac{1}{2} \times (\text{sum of parallel sides}) \times \text{height}$.
        \end{itemize}
    \end{enumerate}

    The height $h$ of the trapezoid can be calculated as follows: $h = 1 \cdot \sin(60^\circ) = \frac{\sqrt{3}}{2}$.

    The area of one trapezoid is: $\text{Area of one trapezoid} = \frac{1}{2} \times (1 + 1) \times \frac{\sqrt{3}}{2} = \frac{1}{2} \times 2 \times \frac{\sqrt{3}}{2} = \frac{\sqrt{3}}{2}$.

    Since there are two such trapezoids, the total area of the hexagon is: $\text{Total Area} = 2 \times \frac{\sqrt{3}}{2} = \sqrt{3}$.

    \vspace{1em}

    \noindent\textbf{Qwen2.5-VL-7B-Instruct:}

    To find the area of the given figure, we need to recognize that the shape can be divided into two congruent trapezoids. Let's break down the problem step by step:

    \begin{enumerate}
        \item \textbf{Identify the Shape}: The given figure is a hexagon with specific angle and side lengths. Since $AF$ is parallel to $CD$, $AB$ is parallel to $EF$, and $BC$ is parallel to $ED$, the hexagon can be divided into two congruent trapezoids by drawing a diagonal from $A$ to $D$.
        \item \textbf{Calculate the Area of One Trapezoid}: Each trapezoid is formed by two right triangles and a rectangle in the middle. The key angles are $60^\circ$ at $FAB$ and $BCD$. This means that the trapezoid can be split into a rectangle and two 30-60-90 triangles.
        \item \textbf{Properties of 30-60-90 Triangles}: In a 30-60-90 triangle, the sides are in the ratio $1 : \sqrt{3} : 2$. Here, the shorter leg (opposite the $30^\circ$ angle) is 1, so the longer leg (opposite the $60^\circ$ angle) is $\sqrt{3}$, and the hypotenuse is 2.
        \item \textbf{Area Calculation}:
        \begin{itemize}
            \item The height of each trapezoid is the same as the shorter leg of the 30-60-90 triangle, which is 1.
            \item The bases of the trapezoid are the sum of the shorter legs of the two 30-60-90 triangles, which is $1 + 1 = 2$.
            \item The area of one trapezoid is given by the formula for the area of a trapezoid: $\text{Area} = \frac{1}{2} \times (\text{Base}_1 + \text{Base}_2) \times \text{Height}$.
            \item Substituting the values, we get: $\text{Area} = \frac{1}{2} \times (1 + 1) \times 1 = \frac{1}{2} \times 2 \times 1 = 1$.
        \end{itemize}
        \item \textbf{Total Area}: Since the hexagon is composed of two such congruent trapezoids, the total area is $1 + 1 = 2$.
    \end{enumerate}

    Therefore, the area of the figure is \boxed{2}.

    \end{mdframed}
\end{figure*} 

\begin{figure*}[p] % [p] 建议改为 p，允许独立成页，或者使用 [t]
    \centering
    \begin{mdframed}[
        frametitle={Reasoning Example 3 (Part 1)},
        frametitlebackgroundcolor=lightyellow,
        backgroundcolor=gray!10,
        linecolor=gray,
        frametitlerule=true,
        frametitlefont=\color{black}\bfseries
    ]
        \centering
        \small
        \includegraphics[width=0.4\linewidth]{img/example/mathvision3.png} 
        \captionof{figure}{\centering Question: The solid shown in the diagram has 12 regular pentagonal faces... (content omitted for brevity) ... Total of the numbers written on the solid?\\Ground Truth: 50} 
        \flushleft

    % --- DMLR 模型部分 ---
    \noindent\textbf{VLAA-Thinker-Qwen2.5VL-7B + DMLR:}

    To solve this problem, we need to determine the number of triangular and square faces on the solid. Given that there are 12 pentagonal faces, each surrounded by 5 square faces, we can calculate the total number of square faces involved in these connections.

    Each pentagonal face is connected to 5 square faces, so for 12 pentagonal faces, the total count of square faces from pentagons alone would be $12 \times 5 = 60$. However, since each square face is shared between two pentagons, the actual number of unique square faces is half of this, which is $60 / 2 = 30$.

    Next, let's denote the number of triangular faces as $T$ and the number of square faces as $S$. We know: $S = 30$.
    The total number of faces is given by the sum of pentagonal, triangular, and square faces: $12 + T + 30 = F$, where $F$ is the total number of faces. Since each triangular face is surrounded by 3 square faces, and each square face is counted twice (once per triangle and once per pentagon), we can use the fact that each square face is part of 3 triangles or 2 pentagons to find $T$.

    Given the structure, we can derive: $3T = 2 \times 30 \implies 3T = 60 \implies T = 20$.

    Thus, the total number of faces $F$ is: $F = 12 + 20 + 30 = 62$.

    Now, calculating the total sum of numbers written on the solid:
    \begin{itemize}
        \item Each pentagonal face contributes $5$: $12 \times 5 = 60$
        \item Each triangular face contributes $1$: $20 \times 1 = 20$
        \item Each square face contributes $-1$: $30 \times -1 = -30$
    \end{itemize}

    Summing these contributions: $60 + 20 - 30 = 50$.

    Therefore, the total of the numbers written on the solid is $50$.

    % 这里结束 Part 1 的框
    \end{mdframed}
\end{figure*}

\begin{figure*}[t] % 第二部分放在下一页顶部
    \centering
    \begin{mdframed}[
        frametitle={Reasoning Example 3 (Part 2)}, % 标题注明是继续
        frametitlebackgroundcolor=lightyellow,
        backgroundcolor=gray!10,
        linecolor=gray,
        frametitlerule=true,
        frametitlefont=\color{black}\bfseries
    ]
    \flushleft
    \small

    % --- Base 模型部分 ---
    \noindent\textbf{VLAA-Thinker-Qwen2.5VL-7B:}

    \texttt{<think>}
    To solve this problem, we need to determine the number of each type of face (pentagon, triangle, and square) on the solid and then sum the values written on these faces according to the given rules.

    \begin{enumerate}
        \item \textbf{Identify the types of faces and their properties}:
        \begin{itemize}
            \item The solid has 12 pentagonal faces.
            \item Let $T$ be the number of triangular faces.
            \item Let $S$ be the number of square faces.
        \end{itemize}
        \item \textbf{Use the given information about the arrangement of faces}:
        \begin{itemize}
            \item Each pentagonal face is surrounded by 5 square faces.
            \item Each triangular face is surrounded by 3 square faces.
        \end{itemize}
        \item \textbf{Set up equations based on the surrounding faces}:
        \begin{itemize}
            \item Since each pentagonal face is surrounded by 5 square faces, and there are 12 pentagonal faces, the total contribution from pentagons to the count of square faces is $12 \times 5 = 60$.
            \item Since each triangular face is surrounded by 3 square faces, the total contribution from triangular faces to the count of square faces is $3T$.
        \end{itemize}
        \item \textbf{Since each square face is counted multiple times (once for each face it touches), we can set up the equation for the total number of square faces}:
        $60 + 3T = 4S$.
        This equation comes from the fact that each square face is part of one pentagonal face and $T$ triangular faces, contributing to the total count of square faces.
        \item \textbf{Determine the number of triangular faces $T$}: We know the total number of faces is $12 + T + S$. However, since each triangular face contributes 3 square faces and each pentagonal face contributes 5 square faces, we can use the relationship derived: $60 + 3T = 4S$.
        \item \textbf{Solve for $S$ in terms of $T$}: $4S = 60 + 3T \implies S = \frac{60 + 3T}{4}$.
        \item \textbf{Substitute back into the total face count equation}: $12 + T + S = 12 + T + \frac{60 + 3T}{4} = 12 + T + 15 + \frac{3T}{4} = 27 + \frac{7T}{4}$.
        \item \textbf{Solve for $T$ such that $S$ is an integer}:
        \begin{itemize}
            \item For $S$ to be an integer, $\frac{60 + 3T}{4}$ must be an integer. Testing values: $60 + 3T \equiv 0 \pmod{4} \implies 3T \equiv -60 \pmod{4} \implies 3T \equiv 0 \pmod{4} \implies T \equiv 0 \pmod{4}$.
            \item Trying $T = 4$: $S = \frac{60 + 3 \times 4}{4} = \frac{72}{4} = 18$.
        \end{itemize}
        \item \textbf{Verify the solution}:
        \begin{itemize}
            \item Total faces: $12 + 4 + 18 = 34$
            \item Square faces: $60 + 3 \times 4 = 72$ divided by 4: $18$
        \end{itemize}
        \item \textbf{Calculate the total value}:
        \begin{itemize}
            \item Pentagons: $12 \times 5 = 60$
            \item Triangles: $4 \times 1 = 4$
            \item Squares: $18 \times (-1) = -18$
            \item Total: $60 + 4 - 18 = 46$
        \end{itemize}
    \end{enumerate}

    Thus, the total of the numbers written on the solid is \boxed{46}.
    \texttt{</think>}

    \noindent\texttt{<answer>} \boxed{46} \texttt{</answer>}

    \end{mdframed}
    % 这一部分不需要 caption，或者可以使用 \captionof{figure}{... continued}
\end{figure*}

\begin{figure*}[p]
    \centering
    \begin{mdframed}[
        frametitle={Reasoning Example 4 (Part 1)},
        frametitlebackgroundcolor=lightyellow,
        backgroundcolor=gray!10,
        linecolor=gray,
        frametitlerule=true,
        frametitlefont=\color{black}\bfseries
    ]
        \centering
        \small
        \includegraphics[width=0.4\linewidth]{img/example/mathvision4.png} 
        \captionof{figure}{\centering Question: We want to paint each square in the grid with the colours P, Q, R and S, so that neighbouring squares always have different colours. (Squares which share the same corner point also count as neighbouring.) Some of the squares are already painted. In which colour(s) could the grey square be painted?\\Options: A. only Q, B. only R, C. only S, D. either R or S, E. it is not possible.\\Ground Truth: D. either R or S} 
        \flushleft

    % --- DMLR 模型部分 ---
    \noindent\textbf{VLAA-Thinker-Qwen2.5VL-7B + DMLR:}

    To determine the color of the grey square, we need to analyze the constraints given by the problem. The grid is filled with the colors P, Q, R, and S such that no two adjacent squares (including diagonally adjacent ones) share the same color.

    \begin{enumerate}
        \item \textbf{Identify Neighboring Squares}:
        \begin{itemize}
            \item The grey square has neighbors in the following positions:
            \item Top-left: Q
            \item Top-right: 
            \item Bottom-left: 
            \item Bottom-right: 
        \end{itemize}
        \item \textbf{Analyze the Top-Left Neighbor}:
        \begin{itemize}
            \item The top-left neighbor is already colored Q.
            \item Therefore, the grey square cannot be Q.
        \end{itemize}
        \item \textbf{Consider Possible Colors for the Grey Square}:
        \begin{itemize}
            \item Since the grey square cannot be Q, it must be either R or S.
            \item We need to check if both R and S are possible without violating the rule that no two adjacent squares can have the same color.
        \end{itemize}
        \item \textbf{Check for Validity of R}:
        \begin{itemize}
            \item If the grey square were R, then the top-right square would need to be a different color from R and Q. This leaves P as the only valid option for the top-right square since it must differ from both R and Q.
        \end{itemize}
        \item \textbf{Check for Validity of S}:
        \begin{itemize}
            \item If the grey square were S, then the bottom-right square would need to be a different color from S and Q. This leaves P as the only valid option for the bottom-right square since it must differ from both S and Q.
        \end{itemize}
        \item \textbf{Conclusion}:
        \begin{itemize}
            \item Both R and S are valid options for the grey square because they do not violate the adjacency rules when considering their neighboring squares.
        \end{itemize}
    \end{enumerate}

    Therefore, the grey square could be painted either R or S.

    \end{mdframed}
\end{figure*}

\begin{figure*}[t]
    \centering
    \begin{mdframed}[
        frametitle={Reasoning Example 4 (Part 2)},
        frametitlebackgroundcolor=lightyellow,
        backgroundcolor=gray!10,
        linecolor=gray,
        frametitlerule=true,
        frametitlefont=\color{black}\bfseries
    ]
        \flushleft
        \small

    % --- Base 模型部分 ---
    \noindent\textbf{VLAA-Thinker-Qwen2.5VL-7B:}

    \texttt{<think>}
    To determine the color of the grey square, we need to consider the constraints given: no two neighboring squares can have the same color, including those that share a corner.

    \begin{enumerate}
        \item The square directly above the grey square is already colored 'Q'.
        \item The square to the left of the grey square is colored 'R'.
        \item The square below the grey square is colored 'Q'.
    \end{enumerate}

    Given these constraints:
    \begin{itemize}
        \item The grey square cannot be 'Q' because it shares a corner with the square above it which is 'Q'.
        \item The grey square cannot be 'R' because it shares a corner with the square to the left which is 'R'.
        \item The grey square cannot be 'P' because there is no information suggesting 'P' is used elsewhere in a way that would conflict with the rules for the grey square.
    \end{itemize}

    Therefore, the only remaining color for the grey square is 'S'.
    \texttt{</think>}

    \noindent\texttt{<answer>} \boxed{S} \texttt{</answer>}

    \end{mdframed}
    \label{fig8}
\end{figure*}

\section{Prompt}
\label{g}
In this section, we provide the complete set of prompts used in our experiments. These prompts cover all evaluation settings, including reasoning, perception, visual-grounding analysis, confidence-based studies, and ablation configurations. By releasing the full prompt formulations, we aim to ensure that every component of our experimental pipeline is transparent and fully reproducible, enabling researchers to faithfully replicate our results and conduct further analysis based on the same standardized inputs.

\label{prompt}
\begin{figure*}[t]
\centering
\begin{tcolorbox}[
    width=\textwidth,            % <<< 跨双栏宽度
    colback=gray!10,
    colframe=gray,
    coltitle=white,
    fonttitle=\bfseries,
    title=Reasoning Chains Correct Judge,
    attach boxed title to top left,
    boxed title style={
        colback=gray!50!black,
        boxrule=0pt,
        arc=0pt,
        outer arc=0pt,
        top=3pt,
        bottom=3pt,
        left=4pt,
        right=4pt
    },
    enhanced,
    sharp corners,
    left=6pt,
    right=6pt,
    top=6pt,
    bottom=6pt,
    halign title=flush left,     % <<< 标题左对齐
    before upper=\raggedright    % <<< 内容左对齐
]
You are given a multimodal reasoning example.

-- The model receives an image and a question (possibly with answer options).\\
-- The model then produces a step-by-step reasoning chain, followed by a final answer.\\
-- You are also given the ground-truth correct answer.\\

Your task is to decide whether the model's final answer is correct or incorrect.

Please follow these rules:\\
1.~Ignore style and verbosity. Focus only on factual correctness.\\
2.~If the final answer matches the ground-truth answer, label it as \textbf{CORRECT}, even if some intermediate reasoning steps are slightly suboptimal.\\
3.~If the final answer does \textbf{NOT} match the ground-truth answer, label it as \textbf{INCORRECT}, even if parts of the reasoning look plausible.\\
4.~If the model outputs multiple conflicting answers, treat the final one as its final answer.\\

Return your decision in the following JSON format only:

\begin{verbatim}
{
  "label": "CORRECT" or "INCORRECT",
  "short_explanation": "<one-sentence>"
}
\end{verbatim}

\noindent QUESTION: \texttt{\{QUESTION\_TEXT\}}\\
OPTIONS (if any): \texttt{\{OPTION\_TEXT\}}\\
GROUND-TRUTH ANSWER: \texttt{\{GT\_ANSWER\}}\\
MODEL REASONING CHAIN: \texttt{\{MODEL\_CHAIN\}}\\
MODEL FINAL ANSWER: \texttt{\{MODEL\_FINAL\_ANSWER\}}

\end{tcolorbox}
\end{figure*}

\begin{figure*}[t]
\centering
\begin{tcolorbox}[
    width=\textwidth,
    colback=gray!10,
    colframe=gray,
    coltitle=white,
    fonttitle=\bfseries,
    title=Faithfulness Reasoning Judge,
    attach boxed title to top left,
    boxed title style={
        colback=gray!50!black,
        boxrule=0pt,
        arc=0pt,
        outer arc=0pt,
        top=3pt,
        bottom=3pt,
        left=4pt,
        right=4pt
    },
    enhanced,
    sharp corners,
    left=6pt,
    right=6pt,
    top=6pt,
    bottom=6pt,
    halign title=flush left,
    before upper=\raggedright
]
You are given a multimodal reasoning example.\par

The model receives:\par
- An image (you can refer to its description below), and\par
- A textual question (possibly with answer options).\par

The model then produces a step-by-step reasoning chain and a final answer.\par

Your task is to judge whether the reasoning chain is FAITHFUL to the information
in the question and image description, or UNFAITHFUL.\par

Definitions:\par
- A FAITHFUL chain explicitly uses relevant information from the question and image
  (or its description), and each step logically contributes to deriving the final answer.\par
- An UNFAITHFUL chain either:\par
  * ignores critical visual/textual details,\par
  * relies on generic templates or guesses,\par
  * contains logically inconsistent steps, or\par
  * gives a correct answer but with reasoning that does not match the actual evidence.\par

Instructions:\par
1. Check whether the reasoning steps are grounded in the given question and image description.\par
2. Check whether the logical steps are coherent and consistent with each other.\par
3. A chain can be UNFAITHFUL even if the final answer happens to be correct by coincidence.\par

Return your judgment in the following JSON format only:\par

\begin{verbatim}
{
  "label": "FAITHFUL" or "UNFAITHFUL",
  "short_explanation": "<one-sentence justification>"
}
\end{verbatim}

QUESTION:\par
\texttt{\{QUESTION\_TEXT\}}\par

IMAGE DESCRIPTION (if available):\par
\texttt{\{IMAGE\_DESCRIPTION\}}\par

MODEL REASONING CHAIN:\par
\texttt{\{MODEL\_CHAIN\}}\par

MODEL FINAL ANSWER:\par
\texttt{\{MODEL\_FINAL\_ANSWER\}}

\end{tcolorbox}
\end{figure*}

\begin{figure*}[t]
\centering
\begin{tcolorbox}[
    width=\textwidth,
    colback=gray!10,
    colframe=gray,
    coltitle=white,
    fonttitle=\bfseries,
    title=Visual Hallucination Judge,
    attach boxed title to top left,
    boxed title style={
        colback=gray!50!black,
        boxrule=0pt,
        arc=0pt,
        outer arc=0pt,
        top=3pt,
        bottom=3pt,
        left=4pt,
        right=4pt
    },
    enhanced,
    sharp corners,
    left=6pt,
    right=6pt,
    top=6pt,
    bottom=6pt,
    halign title=flush left,
    before upper=\raggedright
]
\small

You are given:\par
- An image,\par
- A question about the image, and\par
- Several reasoning statements produced by a model that refer to visual content.\par

Your task is to check whether each visual statement is supported by the image
(or its description), or whether it hallucinates objects, attributes, or relations
that do not actually appear in the image.\par

Definitions:\par
- A NON-HALLUCINATED (grounded) statement is consistent with the image:
  mentioned objects, attributes, and relations are all present or at least plausible.\par
- A HALLUCINATED statement mentions objects, attributes, numbers, or spatial relations
  that contradict the image or clearly do not appear in it.\par

Instructions:\par
1. Carefully compare each visual statement with the image (or its description).\par
2. For each statement, decide whether it is HALLUCINATED or NON-HALLUCINATED.\par
3. Ignore minor wording differences; focus on factual consistency.\par

Return your output in the following JSON format:\par

\begin{verbatim}
{
  "results": [
    {
      "statement_id": 1,
      "label": "HALLUCINATED" or "NON-HALLUCINATED",
      "short_explanation": "<short reason>"
    },
    {
      "statement_id": 2,
      "label": "...",
      "short_explanation": "..."
    }
    ...
  ]
}
\end{verbatim}

--------------------\par
QUESTION:\par
\texttt{\{QUESTION\_TEXT\}}\par

IMAGE DESCRIPTION (if available):\par
\texttt{\{IMAGE\_DESCRIPTION\}}\par

VISUAL STATEMENTS (from the model's reasoning):\par
1.~\texttt{\{STATEMENT\_1\}}\par
2.~\texttt{\{STATEMENT\_2\}}\par
3.~\texttt{\{STATEMENT\_3\}}\par
...\par

\end{tcolorbox}
\end{figure*}

\begin{figure*}[t]
\centering
\begin{tcolorbox}[
    width=\textwidth,
    colback=gray!10,
    colframe=gray,
    coltitle=white,
    fonttitle=\bfseries,
    title=Reasoning System Prompt (Ablation Study),
    attach boxed title to top left,
    boxed title style={
        colback=gray!50!black,
        boxrule=0pt,
        arc=0pt,
        outer arc=0pt,
        top=3pt,
        bottom=3pt,
        left=4pt,
        right=4pt
    },
    enhanced,
    sharp corners,
    left=6pt,
    right=6pt,
    top=6pt,
    bottom=6pt,
    halign title=flush left,
    before upper=\raggedright
]
\small

A conversation between a User and an Assistant.  
The User asks a question about an image, and the Assistant solves it.  
The Assistant first thinks through the reasoning process internally and then provides the final answer.

The reasoning process and the final answer must be enclosed within  
\texttt{<think>} \texttt{</think>} and \texttt{<answer>} \texttt{</answer>} tags, respectively.  
For example:\par
\texttt{<think>} reasoning process here \texttt{</think>}  
\texttt{<answer>} answer here \texttt{</answer>} \par

You are given:\par
- An image,\par
- A question about the image.\par

Your task:\par
1. Carefully analyze the image.\par
2. Identify objects, attributes, and spatial relations relevant to the question.\par
3. Perform step-by-step visual reasoning.\par
4. Give a concise and correct final answer.\par

Follow these rules:\par
- The reasoning inside \texttt{<think>} should be detailed and explicit.\par
- The final answer inside \texttt{<answer>} should be short and definitive.\par
- Do not mention the image analysis process outside of the \texttt{<think>} block.\par

--------------------\par
QUESTION:\par
\texttt{\{QUESTION\_TEXT\}}\par

IMAGE:\par
\texttt{\{IMAGE\_PLACEHOLDER\}}\par

\end{tcolorbox}
\end{figure*}

\begin{figure*}[t]
\centering
\begin{tcolorbox}[
    width=\textwidth,
    colback=gray!10,
    colframe=gray,
    coltitle=white,
    fonttitle=\bfseries,
    title=Fine-Grained Reasoning Prompt (visualization),
    attach boxed title to top left,
    boxed title style={
        colback=gray!50!black,
        boxrule=0pt,
        arc=0pt,
        outer arc=0pt,
        top=3pt,
        bottom=3pt,
        left=4pt,
        right=4pt
    },
    enhanced,
    sharp corners,
    left=6pt,
    right=6pt,
    top=6pt,
    bottom=6pt,
    halign title=flush left,
    before upper=\raggedright
]
\small

You are an Assistant that performs \textbf{fine-grained visual reasoning}.  
You are given an image and a question. Your task is to examine the image carefully,
identify all visually relevant details, and use them to answer the question.

Your reasoning process must:\par
- Attend to \textbf{small visual cues} such as colors, textures, shapes, numbers, and spatial relations.\par
- Explicitly verify each visual observation before relying on it.\par
- Avoid hallucinating any object, attribute, or relation not present in the image.\par
- Connect visual details step by step to form a coherent reasoning chain.\par

The reasoning process must be written inside \texttt{<think>} \texttt{</think>}.\par
The final short answer must be written inside \texttt{<answer>} \texttt{</answer>}.\par

Example format:\par
\texttt{<think>} detailed visual reasoning here \texttt{</think>}\par
\texttt{<answer>} final answer here \texttt{</answer>}\par

Instructions:\par
1. Begin by describing all key visual details relevant to the question.\par
2. Compare objects, count elements, check colors, sizes, and spatial layouts.\par
3. Use concrete grounding from the image before making any inference.\par
4. Produce the final answer only after completing the visual reasoning.\par

--------------------\par
QUESTION:\par
\texttt{\{QUESTION\_TEXT\}}\par

IMAGE:\par
\texttt{\{IMAGE\_PLACEHOLDER\}}\par

\end{tcolorbox}
\end{figure*}

\twocolumn

\clearpage

{
    \small
    \bibliographystyle{ieeenat_fullname}
    \bibliography{ref}
}
